%% file: main.tex
\documentclass[twocolumn]{article}
\usepackage{arxiv}

\input{preamble}

\usepackage{hyperref}
\usepackage{url}  
\usepackage[nameinlink]{cleveref}

\hypersetup{
	colorlinks=true,
	linkcolor=darkblue,
	filecolor=darkblue,
	urlcolor=darkblue,
	citecolor=darkblue,
	pdfauthor={L. Elsem\"{u}ller, M. Schnuerch, P.-C. B\"{u}rkner, S. T. Radev},
	pdftitle={A Deep Learning Method for Comparing Bayesian Hierarchical Models}
}

\title{A Deep Learning Method for Comparing Bayesian Hierarchical Models}
\author{
    Lasse Elsemüller\\
    Institute of Psychology\\
    Heidelberg University\\
    {\small\texttt{lasse.elsemueller@gmail.com}}
    \And
    Martin Schnuerch\\
    Department of Psychology\\
    University of Mannheim\\
    {\small\texttt{martin.schnuerch@gmail.com}}
    \And
    Paul-Christian Bürkner\\
    Department of Statistics\\
    TU Dortmund University\\
    {\small\texttt{paul.buerkner@gmail.com}}
    \And
    Stefan T. Radev\\
    Cluster of Excellence STRUCTURES\\
    Heidelberg University\\
    {\small\texttt{stefan.radev93@gmail.com}}
}

\begin{document}
\twocolumn[{%
  \begin{@twocolumnfalse}
    \maketitle
  \end{@twocolumnfalse}
}]

\begin{abstract}
Bayesian model comparison (BMC) offers a principled approach for assessing the relative merits of competing computational models and propagating uncertainty into model selection decisions.
However, BMC is often intractable for the popular class of hierarchical models due to their high-dimensional nested parameter structure. 
To address this intractability, we propose a deep learning method for performing BMC on any set of hierarchical models which can be instantiated as probabilistic programs.
Since our method enables \textit{amortized inference}, it allows efficient re-estimation of posterior model probabilities and fast performance validation prior to any real-data application.
In a series of extensive validation studies, we benchmark the performance of our method against the state-of-the-art bridge sampling method and demonstrate excellent amortized inference across all BMC settings.
We then showcase our method by comparing four hierarchical evidence accumulation models that have previously been deemed intractable for BMC due to partly implicit likelihoods.
Additionally, we demonstrate how transfer learning can be leveraged to enhance training efficiency.
We provide reproducible code for all analyses and an open-source implementation of our method.
\end{abstract}

\section{Introduction}
\label{sec:intro}

Hierarchical or multilevel models (HMs) play an increasingly important methodological role in the social and cognitive sciences \autocite{farrell2018computational, rouder2017bayesian}.
HMs embody probabilistic and structural information about nested data occurring frequently in various settings, such as educational research \autocite{ulitzsch2020multiprocess}, experimental psychology \autocite{vandekerckhove2011hierarchical}, epidemiology \autocite{jalilian2021hierarchical} or astrophysics \autocite{hinton2019steve}, to name just a few.
Crucially, HMs can often extract more information from rich data structures than their non-hierarchical counterparts (e.g., aggregate analyses), while retaining a relatively high intrinsic interpretability of their structural components (i.e., parameters).
Moreover, viewed as formal instantiations of scientific hypotheses, HMs can be employed to systematically assign preferences to these hypotheses by means of formal model comparison.
For example, \textcite{haaf2017developing} proposed a powerful framework based on Bayesian HMs for formulating and testing competing theoretical positions on quantitative vs. qualitative individual differences.

We consider Bayesian model comparison (BMC) as a principled framework for comparing and ranking competing HMs via Occam's razor \autocite{kass1995bayes, mackay2003information, lotfi2022bayesian}. 
However, standard BMC is analytically intractable for non-trivial HMs, as it requires marginalization over high-dimensional parameter spaces.
Moreover, BMC for complex HMs without explicit likelihoods (i.e., HMs available only as randomized simulators) becomes increasingly hopeless and precludes many interesting applications in the rapidly expanding field of simulation-based inference \autocite{cranmer2020frontier}.

In this work, we propose to tackle the problem of BMC for arbitrarily complex HMs from a simulation-based perspective using deep learning.
In particular, we build on the BayesFlow framework \autocite{radev2020bayesflow, radev2021amortized} for simulation-based Bayesian inference and propose a novel hierarchical neural network architecture for approximating Bayes factors (BFs) and posterior model probabilities (PMPs) for any collection of HMs.

Our neural approach circumvents the steps of explicitly fitting all models and marginalizing over the parameter space of each model.
Thus, it is applicable to both HMs with explicit likelihood functions and HMs accessible only through Monte Carlo simulations (i.e., with implicit likelihood functions).
Moreover, our neural networks come with an efficient way to compute their calibration error \autocite{guo2017calibration}, which provides an important diagnostic for self-consistency.
Lastly, trained networks can be adapted to related tasks, substantially reducing the computational burden when dealing with demanding simulators.

The remainder of this paper is organized as follows. In \textbf{Section}~\ref{sec:theory}, we introduce the theoretical background and related work on (hierarchical) BMC. 
We then present the rationale and details of our deep learning method in \textbf{Section}~\ref{sec:method}.
In \textbf{Sections}~\ref{subsec:val1} and \ref{subsec:val2}, we present two validation studies of the proposed method: One that includes toy models for illustrative purposes and one that includes two popular classes of models from the field of cognitive psychology.
In \textbf{Section}~\ref{subsec:application}, we then apply our method to compare hierarchical diffusion decision models with partly intractable likelihoods on a real data set.  
Finally, \textbf{Section}~\ref{sec:discussion} summarizes our contributions and discusses future perspectives.

\section{Theoretical Background}
\label{sec:theory}

\subsection{Bayesian Hierarchical Modeling}

In order to streamline statistical analyses, researchers rely on assumptions about the probabilistic structure or symmetry of the assumed data-generating process. 
For instance, the canonical IID assumption in psychological modeling states that (multivariate) observations are independent of each other and sampled from the same latent probability distribution \autocite{singmann2019mixed, nicenboim2022introduction}.

However, more complex dependencies may arise in a variety of contexts. 
For instance, if there are repeated measurements per participant or participants belong to different natural groups (e.g., school classes, working groups), the respective observations exhibit higher correlations within those clusters than across them. 
Ignoring this nested structure in statistical analyses may result in biased conclusions \autocite{singmann2019mixed}.
Bayesian HMs formalize this structural knowledge by assuming that observations are sampled from a multilevel generative process \autocite{gelman2006multilevel}.

For instance, the generative recipe for a \textit{two-level} Bayesian HM can be written as:
\begin{align}\label{eq:two_level}
    \etab &\sim p(\etab)\\
    \thetab_m &\sim p(\thetab\,\given \etab) \text{ for } m=1,\dots,M\\
    \xb_{mn} &\sim p(\xb\,\given\,\thetab_m) \text{ for } n=1,\dots,N_m,
\end{align}
where $\etab$ denotes the group-level parameters, $\thetab_m$ denotes the individual parameters in group $m$ and $\xb_{mn}$ represents the $n$-th observation in group $m$.
Such a model suggests the following (non-unique) factorization of the joint distribution:
\begin{multline}\label{eq:two_level_fact}
    p(\etab, \{\thetab_m\}, \{\xb_{mn}\}) = \\ p(\etab)\prod_{m=1}^M p\left(\thetab_m\, \vline \, \etab \right) \prod_{n=1}^{N_m}p\left(\xb_{mn} \given \thetab\right).
\end{multline}
The set notation $\{\thetab_m\}$ and $\{\xb_{mn}\}$ implies that the number of groups and observations in each group can vary across simulations, data sets and experiments and that these quantities are exchangeable.

HMs can be considered as a compromise between a separate analysis of each group (\textit{no-pooling}) that neglects the information contained in the rest of the data and an aggregate analysis of the data (\textit{complete pooling}) that loses the distinction between intra-group and inter-group variability \autocite{hox2017multilevel}.
The \textit{partial pooling} of information induced by HMs leads to more stable and accurate individual estimates through the \textit{shrinkage} properties of multilevel priors, whereby single estimates inform each other \autocite{gelman2006multilevel, burkner2017brms}. 

\begin{figure*}
    \centering
    \begin{subfigure}{0.49\textwidth}
  	\includegraphics[width=\textwidth]{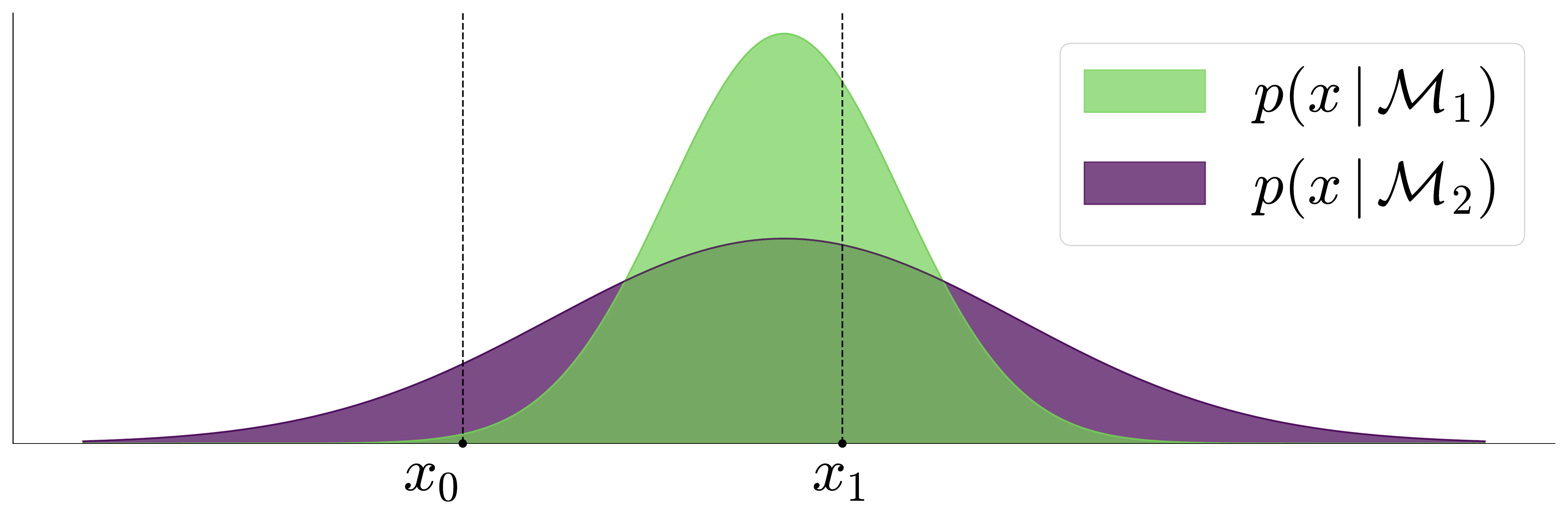}
  	\subcaption{Marginal Likelihoods}
  \end{subfigure}
   \begin{subfigure}{0.49\textwidth}
  	\includegraphics[width=\textwidth]{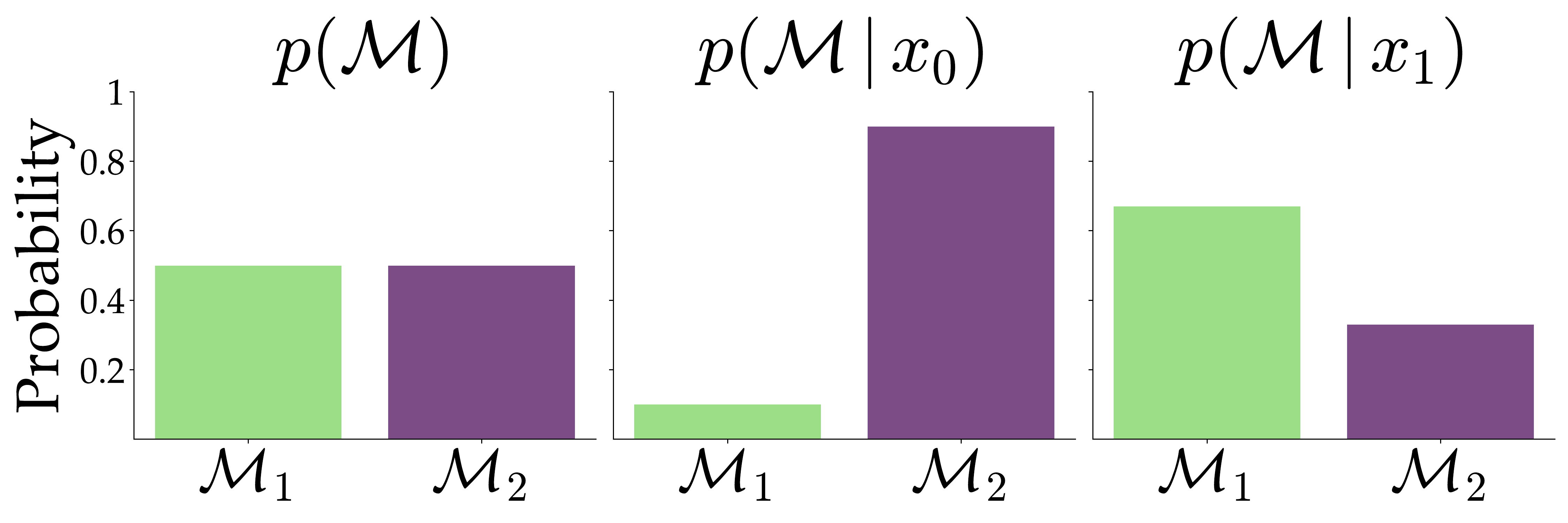}
  	\subcaption{Posterior Model Probabilities}
  \end{subfigure}
  \caption{Hypothetical BMC setting with a simple model $\mathcal{M}_1$ and a more complex model $\mathcal{M}_2$. (a) The complex model which accounts for a broader range of observations needs to spread its marginal likelihood to cover its larger generative scope. It does so at the cost of diminished sharpness. Thus, even though observation $x_1$ is well within its generative scope, the simpler model $\mathcal{M}_1$ yields a higher marginal likelihood and is therefore preferred. In contrast, observation $x_0$ has a higher marginal likelihood under model $\mathcal{M}_2$, as it is very unlikely to be generated by the simpler model $\mathcal{M}_1$. (b) The corresponding posterior model probabilities (PMPs) given a uniform model prior.}\label{fig:bmc}
\end{figure*}

Despite having desirable properties, hierarchical modeling comes at the cost of increased complexity and computational demands.
These increased demands make it hard or even impossible to compare competing HMs within the probabilistic framework of BMC. 
Before we highlight these challenges, we first describe the basics of BMC for non-hierarchical models.

\subsection{Bayesian Model Comparison}
\label{subsec:bmc}

The starting point of BMC is a collection of $J$ competing generative models $\mathcal{M} = \{\mathcal{M}_1,\mathcal{M}_2,\dots, \mathcal{M}_J\}$. 
Each $\mathcal{M}_j$ is associated with a prior $\prior$ on the parameters $\thetab_j$ and a generative mechanism, which is either defined analytically through a (tractable) likelihood density function $\lik$ or realized as a Monte Carlo simulation program $g_j(\thetab, \zb)$ with random states $\zb$.
Together, the prior and the likelihood define the Bayesian joint model 
\begin{align}\label{eq:joint}
    \joint = \prior \lik,
\end{align}
which is also tacitly defined for simulator-based models by marginalizing the joint distribution $p\left(\xb, \zb\given\thetab_j, \model\right)$ over all possible execution paths (i.e., random states) of the simulation program to obtain the implicit likelihood
\begin{align}
    \lik = \int p\left(\xb, \zb \given \thetab_j, \model \right)\text{d}\zb.
\end{align}
This integral is typically intractable for complex simulators \autocite{cranmer2020frontier}, which makes it impossible to evaluate the likelihood and use standard Bayesian methods for parameter inference or model comparison.

The likelihood function, be it explicit or implicit, is a key object in Bayesian inference.
When the parameters $\thetab$ are systematically varied and the data $\xb$ held constant, the likelihood quantifies the relative fit of each model instantiation (defined by a fixed configuration $\thetab$) to the observed data. 

When we marginalize the Bayesian joint model (Eq.~\ref{eq:joint}) over its parameter space, we obtain the \textit{marginal likelihood} or \textit{Bayesian evidence} \autocite[see][Chapter~28]{mackay2003information}:
\begin{align}\label{eq:marg}
  p\left(\xb \given \model \right) = \int \lik \prior \text{d}\thetab_j .
\end{align}
The marginal likelihood can be interpreted as the probability that we would generate data $\xb$ from model $\model$ when we randomly sample from the model's parameter prior $\prior$.
Moreover, the marginal likelihood is a central quantity for prior predictive hypothesis testing or model selection \autocite{kass1995bayes, o1995fractional, rouder2012default}.
It is well-known that the marginal likelihood encodes a notion of Occam’s razor arising from the basic principles of probability \autocite[][see also \autoref{fig:bmc}]{kass1995bayes}. 
Thus, the marginal likelihood provides a foundation for the widespread use of Bayes factors \autocite[BFs;][]{heck2022review} or posterior model probabilities \autocite[PMPs;][]{congdon2006bayesian} for BMC.

The relative evidence for a pair of models can be computed through the ratio of marginal likelihoods for the two competing models $\model$ and $\mathcal{M}_k$,
\begin{align}\label{bf}
    \text{BF}_{jk} = \frac{p\left(\xb\given\model \right)}{p\left(\xb \given \modelk \right)}.
\end{align}
This ratio is called \textit{Bayes factor} (BF) and is widely used for quantifying pairwise model preference in Bayesian settings \autocite{kass1995bayes, heck2022review}.
Accordingly, a $\text{BF}_{jk}$ $> 1$ indicates preference for model $j$ over model $k$ given available data $\xb$.
Alternatively, one can directly focus on the (marginal) posterior probability of a model $\model$,
\begin{equation}\label{eq:pmp}
    p\left(\model\given\xb\right) = \frac{p(\xb\given \model)\,p(\model)}{\sum_{j=1}^J p\left(\xb\given \model\right)\,p(\model)},
\end{equation}
where $p(\model)$ is a categorical (typically uniform) prior distribution encoding a researcher's prior beliefs regarding the plausibility of each considered model.
This prior distribution is then updated with the information contained in the marginal likelihood $p(\xb\given \model)$ to obtain the corresponding \textit{posterior model probability} (PMP), $p(\model\given\xb)$.
Occasionally in the text, we will refer to the vector of PMPs for all $J$ models as $\bs{\pi}$ and to the individual PMPs as $\pi_j$.
The ratio of two PMPs, known as \textit{posterior odds}, is in turn connected to the Bayes factor via the corresponding model priors:
\begin{equation}\label{eq:postodds}
    \frac{p(\model\given\xb)}{p(\modelk\given\xb)} = \frac{p\left(\xb\given\model \right)}{p\left(\xb\given\modelk \right)} \times \frac{p(\model)}{p(\modelk)}.
\end{equation}

Despite its intuitive appeal, the marginal likelihood (and thus BFs and PMPs) represents a well-known and widely appreciated source of intractability in Bayesian workflows, since it typically involves a multi-dimensional integral (Eq.~\ref{eq:marg}) over potentially unbounded parameter spaces \autocite{lotfi2022bayesian, gronau2017tutorial}. 
Furthermore, the marginal likelihood becomes doubly intractable when the likelihood function is itself not available (e.g., in simulation-based settings), thereby making the comparison of such models a challenging and sometimes, up to this point, hopeless endeavor. 

Unsurprisingly, estimating the marginal likelihood (Eq.~\ref{eq:marg}) in the context of hierarchical models becomes even more challenging, since the number of parameters over which we need to perform marginalization grows dramatically (i.e., parameters at all hierarchical levels enter the computation). 
These computational demands render the probabilistic comparison of HMs based on BFs or PMPs analytically intractable even for relatively simple models with explicit (analytical) likelihoods.  
Therefore, researchers need to resort to costly, approximate methods which typically only work for models with explicit likelihoods \autocite{gelman1998simulating, meng2002warp, gronau2017tutorial}.

\subsection{Approximate Bayesian Model Comparison}

\subsubsection{Explicit Likelihoods}
The most efficient approximate methods to date require all candidate models to possess explicitly available likelihood functions. 
For the most simple scenario in which two HMs are nested (e.g., through an equality constraint on a parameter), the Savage-Dickey density ratio \autocite{dickey1970weighted} provides a convenient approximation of the BF \autocite{wagenmakers2010bayesian}.
Typically, however, the candidate models are not nested but exhibit notable structural differences.
Thus, a general-purpose method is needed to encompass the entire plethora of model comparison scenarios arising in practical applications.

A more general method, and the current state-of-the-art for comparing HMs in psychological and cognitive modeling \autocite{gronau2019simple, gronau2020computing, schad2022workflow}, is given by bridge sampling \autocite{bennett1976efficient, meng1996simulating}. 
Bridge sampling has enabled comparisons within families of complex process models, such as multinomial processing trees \autocite[MPTs;][]{gronau2019simple} or evidence accumulation models \autocite[EAMs;][]{gronau2020computing}, and serves as a simple add-on for Markov chain Monte Carlo (MCMC) based Bayesian workflows.

Crucially, bridge sampling relies on the posterior draws generated by an MCMC sampler \autocite[e.g., Stan;][]{carpenter2017stan} to efficiently approximate the marginal likelihood of each respective model \autocite{gronau2017tutorial}.
Note, however, that bridge sampling requires considerably more random draws for stable results than standard parameter estimation \autocite[usually about an order of magnitude more;][]{gronau2017Rbridgesampling}.
Moreover, the approximation quality of bridge sampling is dependent on the convergence of the MCMC chains \autocite{gronau2020computing}. 
Finally, there are no strong theoretical guarantees that the approximations are unbiased and accurately reflect the true marginal likelihoods \autocite{schad2022workflow}.

\subsubsection{Implicit Likelihoods}

With the rise of complex, high-resolution models, intractable likelihood functions (i.e., functions that do not admit a closed form or are too costly to evaluate) become more and more common in statistical modeling.
Such models are not limited to psychology and cognitive science \autocite{van2019cognition, nicenboim2022introduction}, but are also common in fields such as neuroscience \autocite{gonccalves2020training}, epidemiology \autocite{radev2021outbreakflow}, population genetics \autocite{pudlo2016reliable} or astrophysics \autocite{hermans2021towards}.
Despite the common term \textit{likelihood-free}, simulator-based models still possess an implicitly defined likelihood (see \textbf{Section}~\ref{subsec:bmc}) from which we can obtain random draws through Monte Carlo simulations.
This enables model comparison through simulation-based methods, usually by means of approximate Bayesian computation \autocite[ABC;][]{pudlo2016reliable, marin2018likelihood, mertens2018abrox}.

Traditional (rejection-based) ABC methods for BMC repeatedly simulate data sets from the specified generative models, retaining only those simulations that are sufficiently similar to the empirical data.
To enable the calculation of this (dis-)similarity even in high-dimensional cases, the information contained in the simulated data sets is reduced by computing hand-crafted summary statistics, such as the mean and variance \autocite{abc1, abc2}. 
The resulting acceptance rates of the candidate models represent the approximations of their PMPs \autocite{marin2018likelihood, mertens2018abrox}.

Even for non-hierarchical models, ABC methods are known to be notoriously inefficient and highly dependent on the concrete choice of summary statistics \autocite{marin2018likelihood, cranmer2020frontier}. 
This choice is even more challenging for HMs, as modelers now have to retain an optimal amount of information on multiple levels.
Moreover, the rapidly growing number of summary statistics reduces the probability that a simulated data set is similar enough to the empirical data, which vastly increases the number of required simulations \autocite{beaumont2010approximate, marin2018likelihood}.

Regardless of the number of summary statistics, their manual computation carries the danger of insufficiently summarizing the simulations and thereby producing biased approximations \autocite[a phenomenon known as \textit{curse of insufficiency};][]{marin2018likelihood}.
While many improvements of rejection-based ABC have been proposed, most notably ABC-MCMC \autocite{marjoram2003markov, turner2014generalized}, ABC-SMC \autocite{sisson2007sequential}, as well as Gibbs ABC \autocite{turner2014hierarchical} for Bayesian hierarchical modeling in particular \autocite[see also][]{clarte2021componentwise, fengler2021likelihood}, these advancements are still limited by their dependence on hand-crafted summary statistics or kernel density estimation methods.

\begin{figure*}
    \centering
    \includegraphics[width=.87\textwidth]{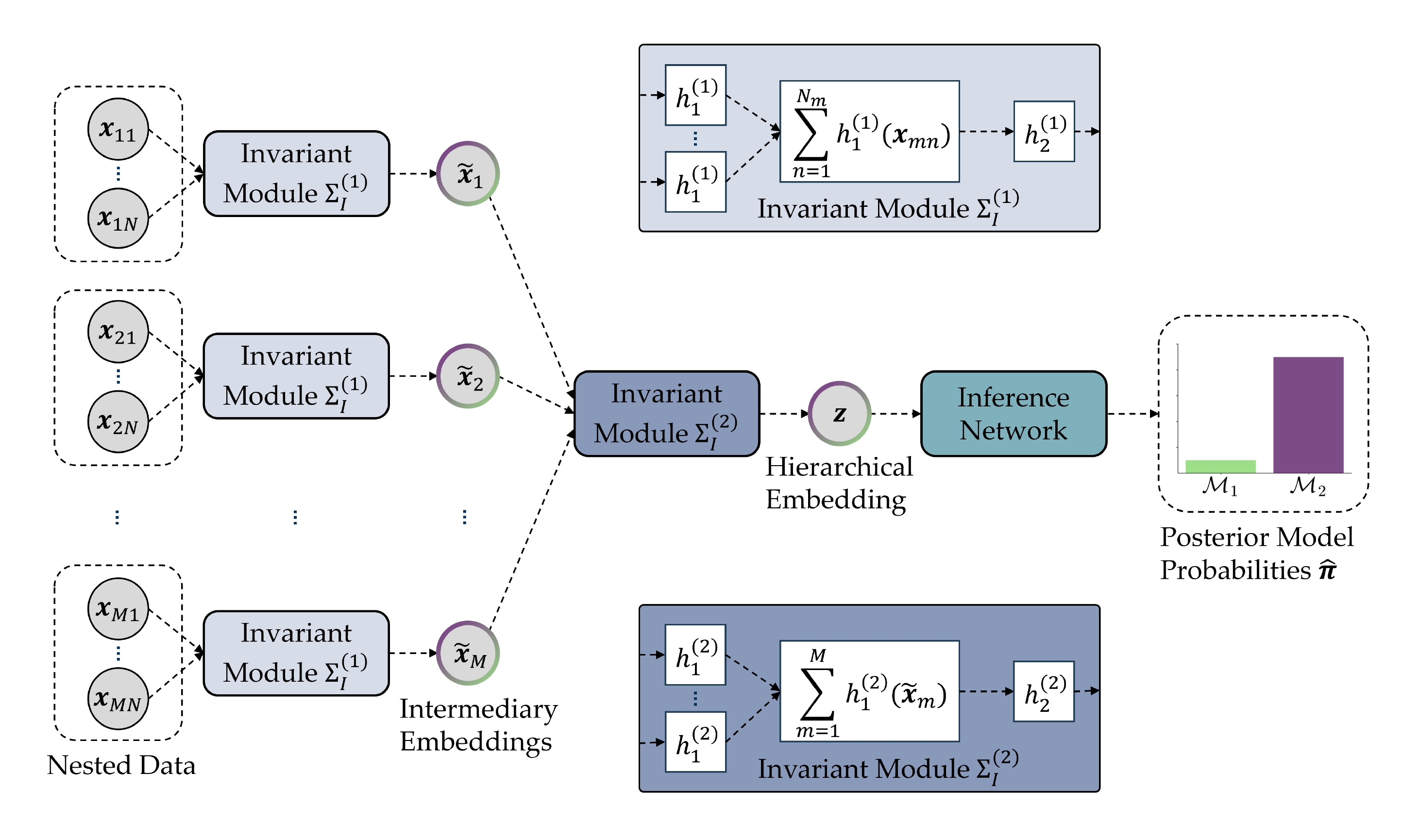}
    \caption{Our proposed hierarchical neural network architecture for encoding permutation invariance in the transformation of nested, two-level data into posterior model probabilities. A first \smash{\textit{invariant module} $\Sigma_I^{(1)}$} reduces all $N_m$ observations within each of the $M$ groups to a single intermediary embedding vector $\widetilde{\xb}_m$. For readability of the figure, we display $N_m = N$ as constant across each group. A second \smash{invariant module $\Sigma_I^{(2)}$} reduces all intermediary embedding vectors to a hierarchical embedding vector $\zb$, which gets passed through an \textit{inference network} to arrive at the final vector $\widehat{\pib}$ of approximated posterior model probabilities.}
    \label{fig:conceptual_hier}
\end{figure*}

Recent developments, such as ABC-RF \autocite{pudlo2016reliable}, combine ABC with machine learning methods to build more expressive approximators for BMC problems.
Accordingly, model comparison is treated as a supervised learning problem -- the simulated data encompasses a training set for a machine learning algorithm that learns to recognize the true generative model from which the data set was simulated.
The machine learning approach reduces the inefficiency problem that haunts rejection-based ABC methods, but does not alleviate the curse of insufficiency \autocite{marin2018likelihood}.

\subsection{Bayesian Model Comparison with Neural Networks}

Recently, \textcite{radev2021amortized} explored a method for simulation-based BMC using specialized neural networks. 
The authors proposed to jointly train two specialized neural networks using Monte Carlo simulations from each candidate model in $\mathcal{M}$: a \textit{summary network} and an \textit{evidential network}. 
The goal of the summary network is to extract \textit{maximally informative} (in the optimal case, \textit{sufficient}) summary statistics from complex data sets.
The goal of the evidential network is to approximate PMPs as accurately as possible and, optionally, to quantify their epistemic uncertainty.

Importantly, simulation-based training of neural networks enables \textit{amortized inference} for both implicit and explicit likelihood models.
Amortization is a property that ensures rapid inference for an arbitrary amount of data sets after a potentially high computational investment for simulation and training \autocite{radev2020bayesflow, radev2021amortized, mestdagh2019prepaid}.
As a consequence, the calibration \autocite{guo2017calibration, talts2018validating} or the inferential adequacy \autocite{schad2021toward, schad2022workflow} of an amortized Bayesian method are embarrassingly easy to validate in practice.

In contrast, non-amortized methods, such as ABC-MCMC \autocite{turner2014generalized} or ABC-SMC \autocite{sisson2007sequential} need to repeat all computations from scratch for each observed data set.
Thereby, it is often infeasible to assess their calibration or inferential adequacy in the pre-data phase of a Bayesian workflow \autocite{gelman2020bayesian}.

Unfortunately, the evidential method proposed by \textcite{radev2021amortized} is not applicable to HMs due to their nested probabilistic structure which cannot be tackled via previous summary networks.
This severely limits the applicability of the method in quantitative research, where hierarchical models have been advocated as a default choice \autocite{mcelreath2020statistical, rouder2017bayesian, lee2011cognitive}.
In the following, we describe how to extend the original method to enable amortized BMC for HMs.

\section{Method}
\label{sec:method}

At its core, our method involves a multilevel permutation invariant neural network which is aligned to the probabilistic symmetry of the underlying HMs (see \autoref{fig:conceptual_hier} for a visualization).
We hold that any method which does not rely on \textit{ad hoc} summary statistics should take this probabilistic symmetry (e.g., exchangeability) into account in order to ensure the structural faithfulness of its approximations.
Moreover, respecting the probabilistic symmetry implied by a generative model cannot only make simulation-based training easier but also suggests a particular architecture for building neural Bayesian approximators.

\subsection{Permutation Invariance}

Permutation invariance is the functional equivalent of the probabilistic notion of exchangeability \autocite{gelman2006multilevel, bloem2020probabilistic}, which roughly states that the order of random variables should not influence their joint probability.

To illustrate this point, consider the model in Eq.~\ref{eq:two_level_fact}, which has two exchangeable levels by design, indexed by $m \in \{1\dots,M\}$ and $n \in \{1,\dots,N_m\}$.
In a setting familiar to social scientists, we might have $M$ individuals, each of whom provides $N_m$ (multivariate) responses on some scale or in repeated trials of an experiment.
Now, suppose that we want to compare a set of HMs $\mathcal{M} = \{\mathcal{M}_1,\dots,\mathcal{M}_J\}$ of the form given by Eq.~\ref{eq:two_level_fact} that might differ in various ways (e.g., different prior/hyperprior assumptions or disparate likelihoods).
Due to the structure of the models, the PMPs $p(\mathcal{M} \given \{\xb_{mn}\})$ depend on neither the ordering of the individuals nor the ordering of their responses (which also holds true for the corresponding BFs).

More precisely, if $\mathbb{S}(\cdot)$
is an arbitrary permutation of an index set, then 
\begin{equation}
    p(\mathcal{M} \given \{\xb_{mn}\}) = p(\mathcal{M} \given \mathbb{S}(\{\xb_{mn}\}))
\end{equation}
for any $\mathbb{S}(\cdot)$ acting on $\{1\dots,M\} \times \{1,\dots,N_1\} \times \dots \times \{1,\dots,N_M\}$ where $\times$ denotes the Cartesian product of two (index) sets.
Note that this notation implies that only permuting each $m$ and permuting each $n$ \textit{within}, but not \textit{across} each group $m$ is allowed.
The property of permutation invariance is immediately obvious from the right-hand side of Eq.~\ref{eq:two_level_fact} that involves two nested products (products being permutation invariant transformations when seen as functions operating on sets).
Naturally, learning permutation invariance directly from data or simulations is hardly feasible with standard neural networks, even for non-nested data.
Indeed, for non-hierarchical generative models, \textcite{radev2021amortized} propose to use composite permutation invariant networks as employed by \textcite{zaheer2017deep}.
In the following section, we generalize this architectural concept to the hierarchical setting.

\subsection{Hierarchical Invariant Neural Network Architecture}

Permutation invariant networks differ from standard feedforward networks in that they can process inputs of different sizes and encode the probabilistic symmetry of the data directly (i.e., remove the need to learn the symmetry implicitly during training by supervised learning alone).

For the purpose of BMC with HMs, we realize a hierarchical permutation invariant function via a stack of \smash{\textit{invariant modules} $\Sigma_I^{(l)}$} for each hierarchical level $l = 1,\dots,L$ of the Bayesian model (see \autoref{fig:conceptual_hier}).
Each invariant module performs an equivariant non-linear transformation \smash{$h^{(l)}_1$} acting on the individual data points, followed by a pooling operator (e.g., sum or max) and a further non-linear transformation $h^{(l)}_2$ acting on the pooled data.

In order to preserve hierarchical symmetry, we apply each \smash{$\Sigma_I^{(l)}$} independently to each nested sequence of data points.
To make this point concrete, consider the two-level model given by Eq.~\ref{eq:two_level_fact} and let data point $\xb_{mn}$ denote the multivariate response of person $m$ in trial $n$ of some data collection experiment.
Accordingly, the first invariant module $\Sigma_I^{(1)}$ operates by reducing the trial data $\{\xb_{n}\}_m$ of each person $m$ to a single person-vector $\widetilde{\xb}_m$ of fixed size:
\begin{equation}
    \widetilde{\xb}_m = \Sigma_I^{(1)}\left( \{\xb_{n}\}_m \right) = h^{(1)}_2\left(\sum_{n=1}^{N_m} h^{(1)}_1(\xb_{mn})\right)\label{eq:inv_simple},
\end{equation}
where $h_1$ and $h_2$ are implemented as simple feedforward neural networks with trainable parameters suppressed for clarity.
The second invariant module $\Sigma_I^{(2)}$ then compresses all person vectors to a final vector $\zb$ of fixed size:
\begin{equation}
    \zb = \Sigma_I^{(2)}\left( \{\xb_{m}\} \right) = h^{(2)}_2\left(\sum_{m=1}^M h^{(2)}_1(\widetilde{\xb}_m)\right).
\end{equation}
In this way, the architecture becomes completely independent of the number of persons $M$ or number of trials per person $N_m$, which could vary arbitrarily across persons.
The vector $\zb$, whose dimensionality represents a tunable hyperparameter, can be interpreted as encoding learned summary statistics for the BMC task at hand (to be discussed shortly). 
Moreover, it is easy to see that $\zb$ is independent of the ordering of persons or the ordering of trials within persons, as necessitated by the model formulation in Eq.~\ref{eq:two_level_fact}.
Thus, the composition $\Sigma_I^{(2)} \circ \Sigma_I^{(1)}(\{\xb_{mn}\})$ reduces a hierarchical data set with two levels to a single vector $\zb$ which respects the probabilistic symmetry implied by the particular hierarchical model formulation.

\subsection{Increasing the Capacity of Invariant Networks}

Encoding an entire hierarchical data set $\{\xb_{mn}\}$ into a single vector $\zb$ forces the composite neural network to perform massive data compression, creating a potential information bottleneck.
For complex generative models, this task can become rather challenging and will depend highly on the representational capacity of the neural network (i.e., its ability to extract informative data set embeddings).
Fortunately, we can enhance the simple architecture described in the preceding paragraph by using insights from \textcite{zaheer2017deep} and \textcite{bloem2020probabilistic}.

In order to increase the capacity of the previously introduced invariant transformation, we can stack together multiple \smash{\textit{equivariant modules} $\Sigma_E^{(l)}$}.
Each equivariant module implements a combination of equivariant and invariant transformations. 
For instance, focusing on our two-level model example (Eq.~\ref{eq:two_level_fact}), the transformations at level $1$ for each person $m$ are now given by:
\begin{align}
    \widetilde{\xb}_m &= h^{(1)}_2\left(\sum_{n=1}^{N_m} h^{(1)}_1(\xb_{mn})\right)\\
    \widetilde{\xb}_{mn} &= h_3^{(1)}(\left[\xb_{mn}, \widetilde{\xb}_m\right]) \text{\quad for \quad} n=1,\dots,N_m,
\end{align}
where $h_3$ is also implemented as a simple feedforward neural network.
In this way, each intermediary output $\widetilde{\xb}_{mn}$ of the equivariant module now contains information from all data points, so the network can learn considerably more flexible transformations.
Moreover, we can stack $K$ equivariant modules followed by an invariant module, in order to obtain a \textit{deep invariant module}, which for the first hierarchical level ($l = 1$) takes the following form:
\begin{equation}
    \widetilde{\xb}_m = (\Sigma_I^{(1)} \circ \Sigma_E^{(K, 1)} \circ \cdot\cdot\cdot \circ \Sigma_E^{(1, 1)})(\{\xb_{n}\}_m).
\end{equation}
Compared to the simple invariant module from Eq.~\ref{eq:inv_simple}, the deep invariant module involves a larger number of computations but allows the network to learn more expressive representations.
Accordingly, the transformation for the second hierarchical level ($l = 2$), which yields the final summary representation $\zb$, is given by:
\begin{equation}
    \zb = (\Sigma_I^{(2)} \circ \Sigma_E^{(K', 2)} \circ \cdot\cdot\cdot \circ \Sigma_E^{(1, 2)})(\{\widetilde{\xb}_m\}),
\end{equation}
where the number of equivariant modules $K'$ for level $2$ can differ from the number of equivariant modules $K$ for level $1$.
In our experiments, reported in \textbf{Section}~\ref{sec:experiments}, we observe a clear advantage of using deep invariant networks over their simple counterparts.
Furthermore, for two-level models, we find that the performance of the networks is largely insensitive to the choice of $K$ or $K'$.

\subsection{Learning the Model Comparison Problem}
\label{subsec:learning_BMC}

In order to get from the learned summary representation $\zb$ to an approximation of the analytic PMPs $\widehat{\pib}$, we apply a final neural classifier (i.e., the inference network) $\mathcal{I}(\bs{z}) = \widehat{\pib}$, as visualized in \autoref{fig:conceptual_hier}.
We deviate from the Dirichlet-based setting in \textcite{radev2021amortized}, since we found that
implementing the inference network as a standard softmax classifier \autocite{grathwohl2019your} provides slightly better calibration and leads to more stable training in the specific context of HMs.

Denoting the entire hierarchical neural network as $f_{\phib}(\{\bs{x}\}) = \widehat{\bs{\pi}}$ and an arbitrary hierarchical data set as $\{\bs{x}\}$, we aim to minimize the expected logarithmic loss
\begin{equation}\label{eq:expected_loss}
   \min_{\phib} \mathbb{E}_{p(\mathcal{M}, \{\bs{x}\})} \left[ -\sum_{j=1}^J \mathbb{I}_{\mathcal{M}_j} \cdot \log f_{\phib}(\{\bs{x}\})_j \right],
\end{equation}
where $\phib$ represents the vector of trainable neural network parameters (e.g., weights and biases), $\mathbb{I}_{\mathcal{M}_j}$ is the indicator function for the ``true'' model. 
The expectation runs over the joint generative (mixture) distribution of all models $p(\mathcal{M}, \{\bs{x}\})$, which we access through Monte Carlo simulations.
Since the logarithmic loss is a \textit{strictly proper loss} \autocite{gneiting2007strictly}, it drives the outputs of $f_{\phib}(\{\bs{x}\})$ to estimate the actual PMPs $p(\mathcal{M} \given \{\bs{x}\})$ as best as possible. 
Thus, perfect convergence in theory guarantees that the network outputs the analytically correct PMPs which asymptotically select the ``true'' model in the closed world or the model that minimizes the Kullback-Leibler divergence to the ``true'' data generating process in the open world \autocite{barron1999consistency}.

In practice, we approximate Eq.~\ref{eq:expected_loss} over a training set of $B$ simulations from the competing HMs.
Each entry $b$ for $b = 1,...,B$ in this training set represents a hierarchical data set $\{\bs{x}^{(b)}\}$ itself along with a corresponding one-hot encoded vector for the ``true'' model index $\mathcal{M}^{(b)}_j$. 
The latter denotes the model from which the data set was generated and serves as the ``ground truth'' for supervised learning.

Similarly to \textcite{radev2021amortized}, our neural method encodes an implicit preference for simpler HMs (i.e., Occam's razor) inherent in all marginal likelihood-based methods \autocite[see][Chapter~28]{mackay2003information}.
Since our simulation-based training approximates an expectation over the marginal likelihoods of all HMs $p(\mathcal{M})\,p(\bs{x} \given \mathcal{M})$, data sets generated by a simpler HM will tend to be more similar compared to those generated by a more complex one (cf. \autoref{fig:bmc}).
Thus, data sets that are plausible under both HMs will be generated more often by the simpler model than by the more complex model.
A sufficiently expressive neural network will capture this behavior by assigning a higher PMP for the simpler model\footnote{Assuming equal prior model probabilities.}, thereby capturing complexity differences arising directly from the generative behavior of the HMs.

Finally, to increase training efficiency when working under a limited simulation budget, we also explore a novel pre-training method inspired by \textit{transfer learning} \autocite{torrey2010transfer, bengio2009curriculum}.
First, we train the networks on data sets with a reduced number of exchangeable units (e.g., reducing the number of observations at level $l=1$).
This procedure accelerates training since it uses fewer simulator calls and the forward pass through the networks becomes cheaper.
In a second step, we generate data with a realistic number of exchangeable units. 
Crucially, since we can use the pre-trained network from step one as a better-than-random initialization, we need considerably fewer simulations than if we trained the network from scratch. 
Indeed, our real-data application in Experiment~\ref{subsec:application} demonstrates the utility of this training method.

\section{Experiments}
\label{sec:experiments}

In this section, we first conduct two simulation studies in which we extensively test the approximation performance of our hierarchical neural method. 
We start with a comparison of two nested toy HMs in \textbf{Section}~\ref{subsec:val1}, followed by a comparison of two complex non-nested HMs of cognition in \textbf{Section}~\ref{subsec:val2}.
For both validation studies, we test our method internally by examining the calibration of the approximated PMPs.
Additionally, we validate our method externally by benchmarking its performance against the current state-of-the-art for comparing HMs, namely, bridge sampling \autocite{gelman1998simulating, gronau2017Rbridgesampling}.
To enable this challenging benchmark, we limit our validation studies to the comparison of models with explicit likelihoods to which bridge sampling is applicable.

Finally, in a real-data application, we use our deep learning method to compare four hierarchical EAMs of response time data in \textbf{Section}~\ref{subsec:application}.
Two of these models have no analytic likelihood, which makes the entire BMC setup intractable with current state-of-the-art methods (e.g., bridge sampling).
Moreover, with this example, we also address the utility of a novel EAM, the Lévy flight model \autocite{voss2019sequential}, that has previously been impossible to investigate directly using Bayesian HMs.

For all experiments, we assume uniform model priors $p(\model)=1/J$.
All computations are performed on a single-GPU machine with an NVIDIA RTX 3070 graphics card and an AMD Ryzen 5 5600X processor. 
The reported computation times are measured as wall-clock times.
Details on the implementation of our neural networks and the employed training procedures are provided in Appendix~\ref{sec:app_NN_training}.
Code for reproducing all results from this paper is freely available at \url{https://github.com/bayesflow-org/Hierarchical-Model-Comparison}.
Additionally, our proposed method is implemented in the BayesFlow Python library for amortized Bayesian workflows \autocite{radev2023bayesflow}.

\subsection{Validation Study 1: Hierarchical Normal Models}
\label{subsec:val1}

In this first experiment, we examine a simple and controllable model comparison setup to examine the behavior of our method under various conditions, before moving on to more complex scenarios.
Inspired by \textcite{Gronau2021example}, we compare two hierarchical normal models \(\modelone\) and \({\modeltwo}\) that share the same hierarchical structure
\begin{align}
\tau^2 &\sim \text{Normal}_+(0, 1)\\
\sigma^2 &\sim \text{Normal}_+(0, 1)\\
\theta_m &\sim \text{Normal}(\mu, \sqrt{\tau^2}) \text{ for } m=1,\dots,M\\
x_{mn} &\sim \text{Normal}(\theta_m, \sqrt{\sigma^2}) \text{ for } n=1,\dots,N_m,
\end{align}
with Normal$_+(\cdot)$ denoting a zero-truncated normal distribution.
The models differ with respect to the parameter \(\mu\) that describes the location of the individual-level parameters \({\theta_m}\):
Whereas \(\modelone\) assumes the location of \({\theta_m}\) to be fixed at $0$, the more flexible \(\modeltwo\) allows for $\mu$ to vary
\begin{align}
    \modelone \text{: } \mu &= 0\\
    \modeltwo \text{: } \mu &\sim \text{Normal}(0, 1).
\end{align}

\subsubsection{Calibration}
\label{sec:val1_calibration}

The most important properties of an approximate inference method are the trustworthiness of its results and, more pragmatically, whether we can diagnose the lack of trustworthiness in a given application.
A useful proxy for trustworthiness is the \textit{calibration} of a probabilistic classifier, which measures how closely the predicted probabilities of outcomes match their true underlying probabilities \autocite{guo2017calibration, schad2022workflow}. 

However, computing the calibration of a BMC procedure is hardly feasible in a non-amortized setting, since it involves applying the method to a large number of simulated data sets.
For bridge sampling, for example, that would imply re-fitting the models via MCMC and running bridge sampling on at least hundreds, if not thousands of simulated data sets.
The calibration of our networks, on the other hand, can be determined almost immediately after training due to their amortization property \autocite{radev2021amortized}. 

In the following experiments, we assess the calibration of our networks visually (via calibration curves) and numerically (via a measure of calibration error).
For generating a calibration curve \autocite{degroot1983comparison, niculescu2005predicting}, we first sort the predicted PMPs $\widehat{\pi}^{(s)}_j$ on $S$ simulated data sets $s = 1, \ldots, S$, which we then partition into $I$ equally spaced probability bins $i = 1, \ldots, I$ (we use $I = 15$ bins for all validation experiments).
For each model $j$ and each bin $i$ containing a set $\mathcal{B}_{ij}$ of predicted model indices, we compute the mean prediction for the model (predicted probability, PP) and the actual fraction of this model being true (true probability, TP) as follows:
\begin{align}
    \text{PP}(\mathcal{B}_{ij}) &:= \frac{1}{\lvert \mathcal{B}_{ij} \rvert} \sum_{b \in \mathcal{B}_{ij}} \widehat{\pi}^{(b)}_j,\\
    \text{TP}(\mathcal{B}_{ij}) &:= \frac{1}{\lvert \mathcal{B}_{ij} \rvert} \sum_{b \in \mathcal{B}_{ij}} \mathbb{I}_{\mathcal{M}^{(b)}_j},
\end{align}
where $\mathbb{I}$ again denotes the indicator function for the ``true model''.
These two quantities varying over the bins form the $X$- and $Y$-axis of a calibration curve. 
A well-calibrated model comparison method with an agreement in each bin (as indicated by a diagonal line) thus yields approximations that reflect the true probabilities of the compared models \autocite{guo2017calibration}. 
We further summarize this information via the Expected Calibration Error \autocite[ECE;][]{naeini2015obtaining} as a single number bounded between $0$ and $1$, which we estimate by averaging the individual deviations between predicted and true probability in each bin:
\begin{equation}\label{eq:ece}
    \widehat{\text{ECE}}_j := \sum_{i=1}^{I} \frac{\lvert \mathcal{B}_{ij} \rvert}{S} \bigg\lvert \text{PP}(\mathcal{B}_{ij}) - \text{TP}(\mathcal{B}_{ij}) \bigg\rvert.
\end{equation}
If follows from Eq.~\ref{eq:ece} that a perfect ECE can be achieved by always predicting indifferent probabilities (e.g., $\widehat{\pi}_1 = \widehat{\pi}_2 = .5$ when comparing two models).
We therefore complement our calibration assessment by measuring the accuracy of recovery, for which we dichotomize the predicted PMPs $\widehat{\pi}^{(s)}_j$ on $S$ simulated data sets into one-vs-rest model predictions \smash{$\widehat{\mathcal{M}}^{(s)}_j$}:

\begin{equation}
    \text{Acc}_j := \frac{1}{S}\, \mathbb{I}_{\widehat{\mathcal{M}}^{(s)}_j = \mathcal{M}^{(s)}_j}. \label{eq:accuracy}
\end{equation}

Thus, in our BMC context, accuracy roughly is to ECE what sharpness is to posterior calibration in Bayesian parameter estimation \autocite{burkner2022some, clarte2022study}.\footnote{We focus on the accuracy since we use a uniform model prior $p(\mathcal{M})$, but other metrics of predictive performance, such as the logarithmic scoring rule, would have been expedient as well.}

\begin{figure*}
    \centering
    \begin{subfigure}{0.47\textwidth}
     \includegraphics[width=\textwidth]{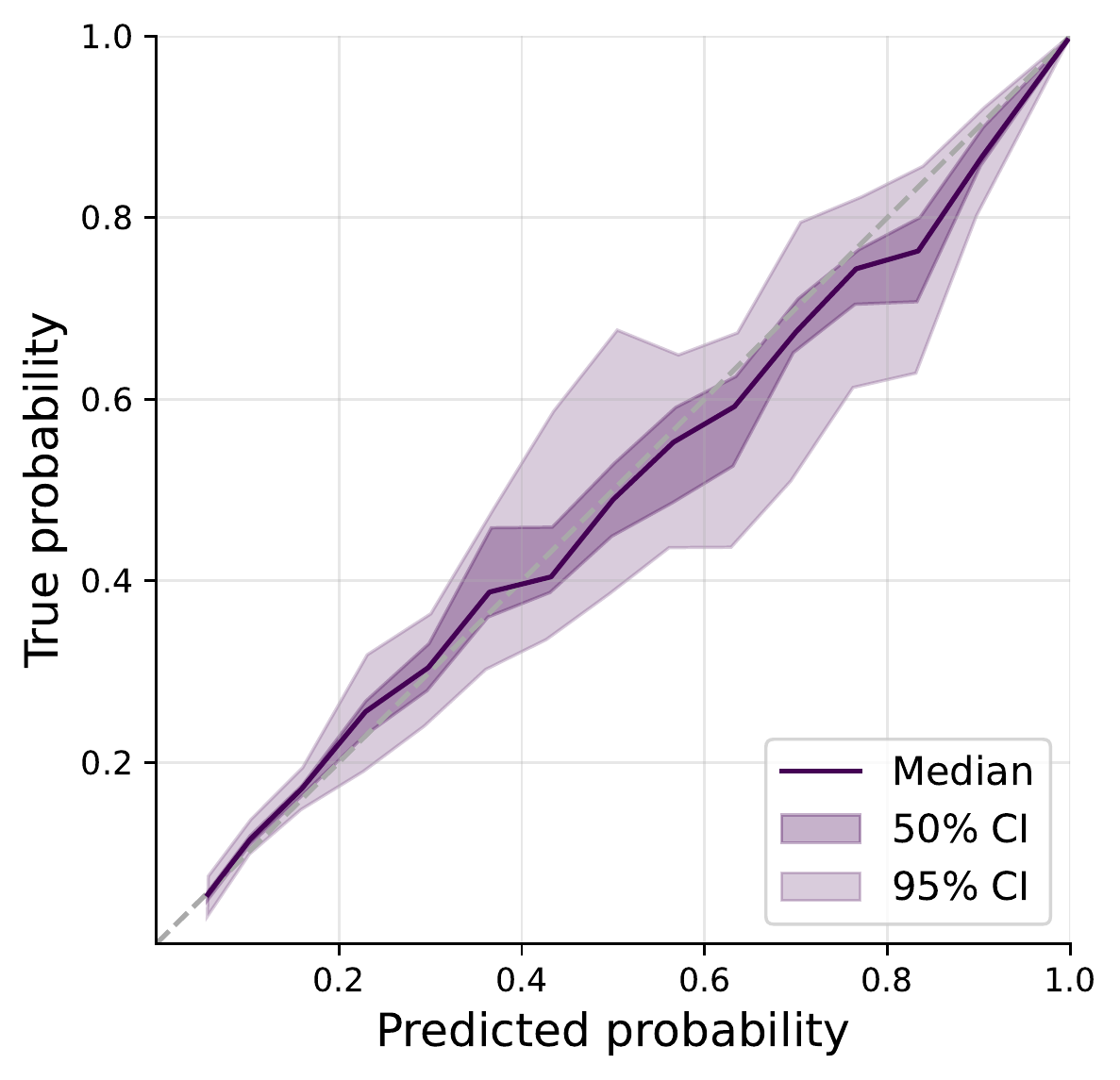}
    \subcaption{Median calibration curve and confidence intervals (CIs) for data sets of $M=50$ groups with $N_m = 50$ observations within each group.}
    \label{fig:cal_curve}
    \end{subfigure}
    \hspace{0.03\textwidth}
    \begin{subfigure}{0.49\textwidth}
     \includegraphics[width=\textwidth]{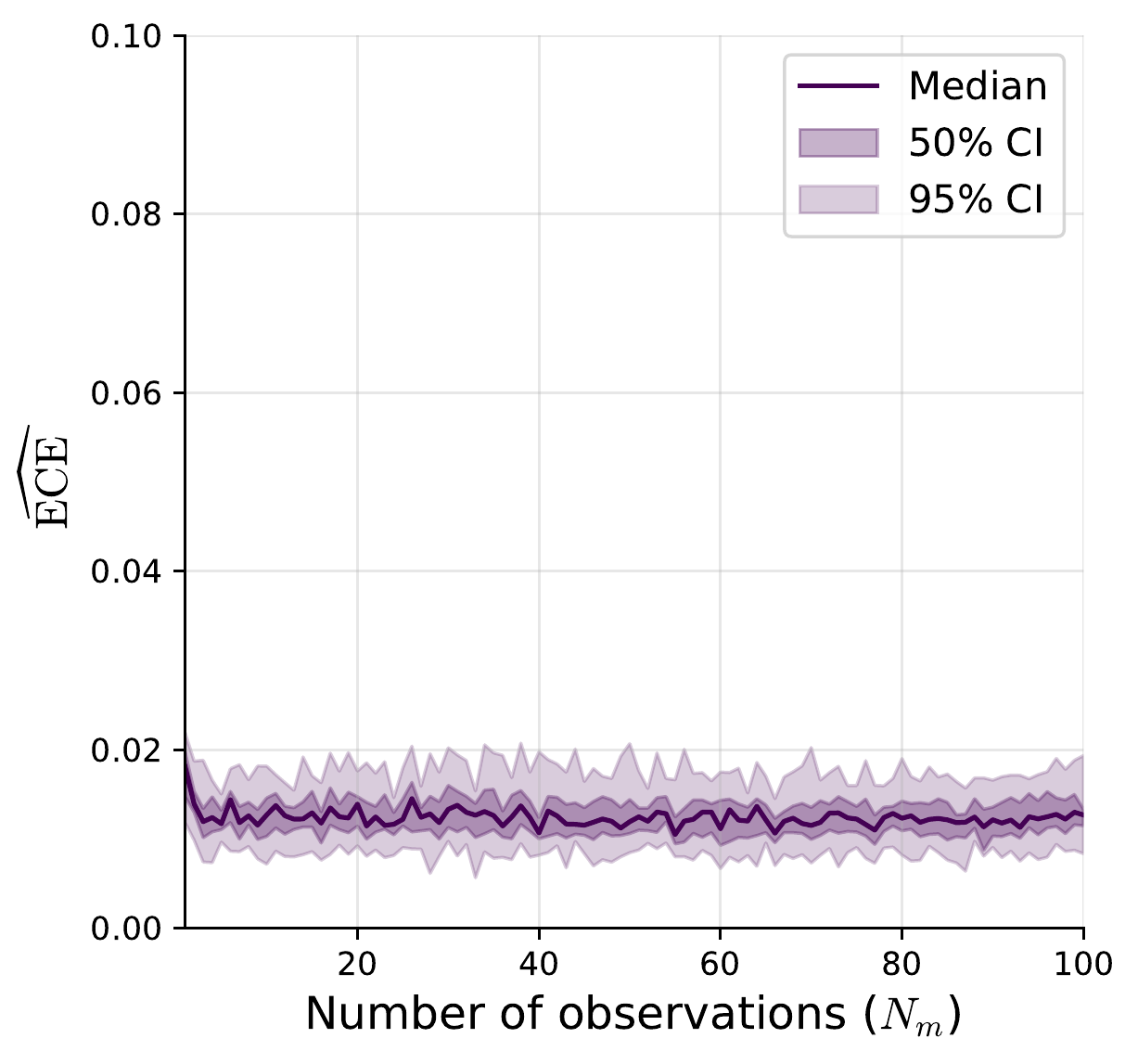}
    \subcaption{Median expected calibration errors (ECEs) and confidence intervals for data sets of $M = 50$ groups with differing numbers of observations $N_m$  within each group.}
    \label{fig:eces_var_obs}
    \end{subfigure}
    \caption{Validation study 1: Calibration results for (a) the neural network trained on fixed data set sizes and (b) the neural network trained on data sets with varying numbers of observations. Medians and confidence intervals (CIs) are computed over $25$ repetitions.}
\end{figure*}

\paragraph{Fixed data set sizes}

In the first calibration experiment, we examine the performance of our method for the most simple application case of learning a model comparison problem on a specific (fixed) data set size. 
Here, all data sets simulated for training and validating the network consist of $M = 50$ groups and $N_m = 50$ observations for each group $m = 1,\dots,M$. 

We train the network for $10,000$ backpropagation steps, taking $6$ minutes.
Subsequently, we calculate its calibration on $5,000$ held-out validation data sets and repeat this process $25$ times to obtain stable results with uncertainty quantification.
\autoref{fig:cal_curve} depicts the resulting median calibration curve. 
Its close alignment to the dashed diagonal line (representing perfect calibration) indicates that the PMP approximations are well-calibrated (median ECE over all repetitions of $\widehat{\text{ECE}} = 0.014$).
The curve's coverage of the full range of predicted probabilities and the median accuracy of $\widehat{\text{Acc}} = .89$ confirm that the excellent calibration does not stem from indifferent predictions.
The subsequent comparison of our method to bridge sampling suggests that this accuracy is indeed close to the upper bound imposed by the aleatoric uncertainty in the model-implied data.

\paragraph{Data sets with varying numbers of observations}

We now train our hierarchical network to approximate BMC over a range of hierarchical data sets with varying numbers of observations within groups $N_m$. 
This amortization over observation sizes would provide a substantial efficiency gain if a researcher desires to compare HMs on multiple data sets with differing $N_m$, as only a single network would have to be trained for all data sets.\footnote{Note that we refer to variability between data sets. We describe an approach for handling within data set variability of nested trials in \textbf{Section}~\ref{subsec:application}.}
In our validation setup, each simulated data set still consists of $M = 50$ groups, but now the number of observations within those groups varies in $N_m = 1,\ldots,100$.

We train the network for $20,000$ training steps, taking $13$ minutes. 
At each training step, we draw the number of observations for the current batch of simulations from a discrete uniform distribution $N_m \sim \text{Uniform}_D(1,100)$.
For each $N_m$ used during training, we evaluate the calibration $25$ times on $5,000$ held-out simulated validation data sets.
This repetition procedure allows us to quantify the uncertainty of our ECE estimates.

\autoref{fig:eces_var_obs} plots the median ECE values for each observation size. 
The neural network achieves high calibration with a median ECE over all observation sizes (and repetitions) of $\widehat{\text{ECE}} = 0.012$.
Moreover, the unsystematic pattern of the median curve and the homoscedastic variation between the observation sizes indicate that the network has learned the model comparison task equally well for all settings (with the ECE only rising slightly for the poorly identifiable $N_m = 1$ setting).
Together, the low calibration error and the accurate model predictions (median accuracy $\widehat{\text{Acc}} = .88$) indicate that our method incurs no trade-off between calibration and accuracy.
We additionally observe no bias towards a model in all but the smallest observation sizes (see \autoref{fig:Acc_SBC_variable_observations} for accuracy and bias examinations in all settings).

\begin{figure*}[h]
    \centering
    \begin{subfigure}{0.49\textwidth}
     \includegraphics[width=\textwidth]{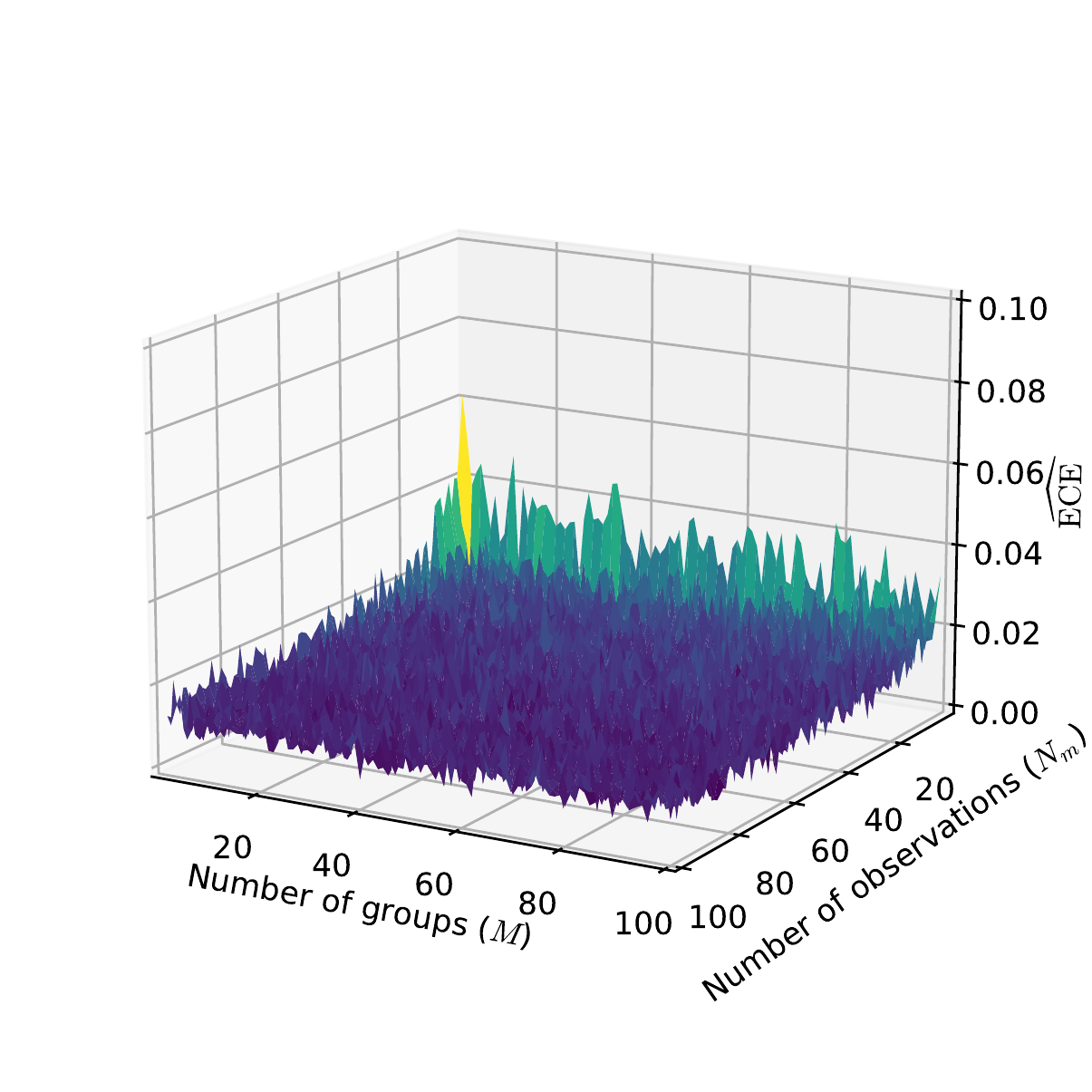}
    \subcaption{Expected Calibration Error (ECE).}
    \label{fig:eces_var_sizes}
    \end{subfigure}
    \begin{subfigure}{0.49\textwidth}
     \includegraphics[width=\textwidth]{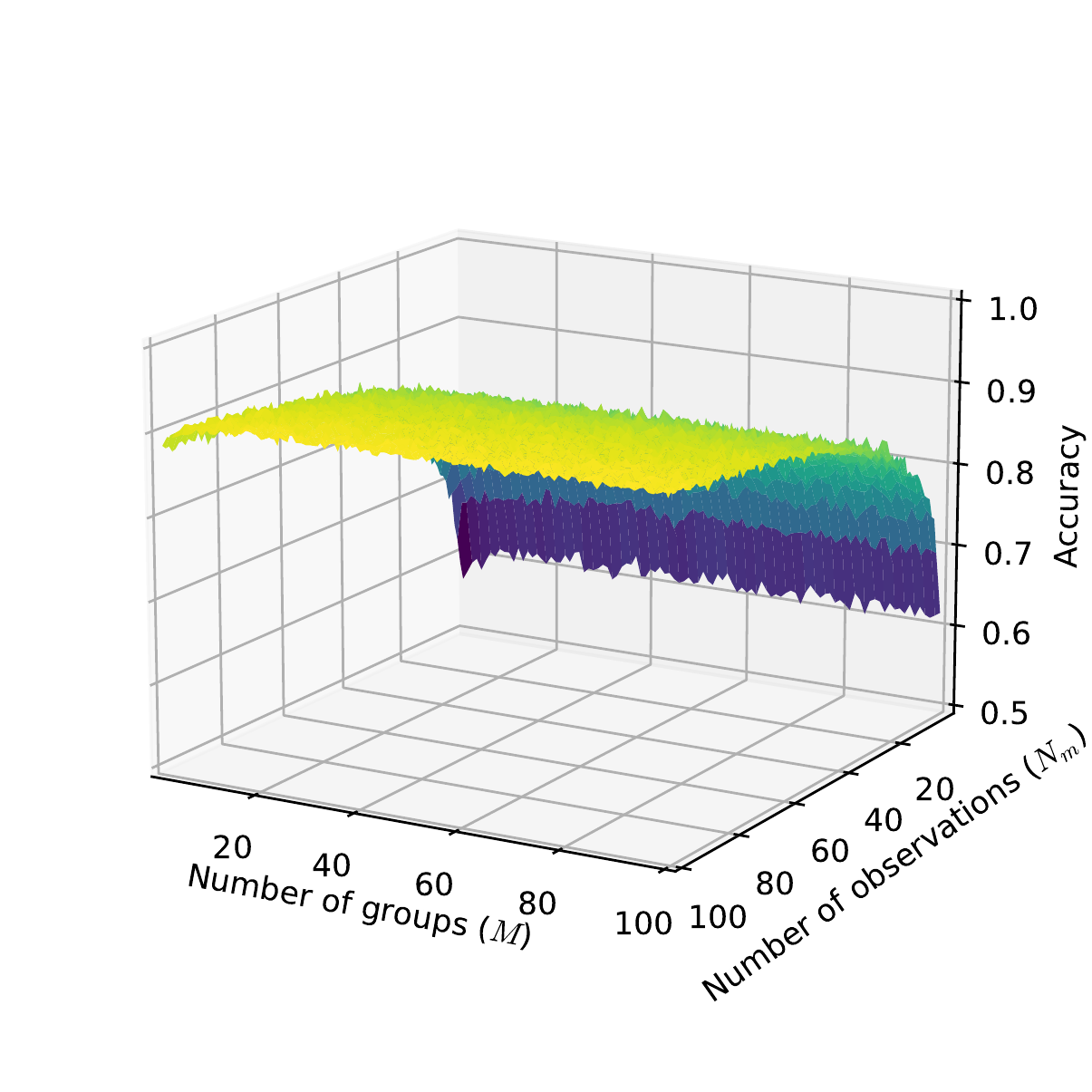}
    \subcaption{Accuracy of recovery.}
    \label{fig:accuracies_var_sizes}
    \end{subfigure}
    \caption{Validation study 1: Results for the neural network trained and tested over variable data set sizes.}
    \label{fig:variable_network_res}
\end{figure*}

\paragraph{Data sets with varying numbers of groups and observations}

In the third calibration experiment, we test the ability of the network to learn a model comparison problem over a range of data sets with varying numbers of groups $M$ and varying observations per group $N_m$. 
This training scheme allows for amortized model comparison on multiple data sets with different sizes, which can be especially useful for a priori sample size determination on simulated data.
Additionally, the trained network can be stored and reused on future data sets with yet-unknown sample sizes. 
For this experiment, training and validation data sets are simulated with $M = 1, \ldots, 100$ groups and $N_m = 1, \ldots, 100$ observations, resulting in a vast variability of data set sizes between $1$ up to $10,000$ data points.

Given the complexity of the learning task, we now train the network for $40,000$ training steps, taking $36$ minutes.
At each training step, we draw the number of groups and observations from discrete uniform distributions $M \sim \text{Uniform}_D(1,100)$ and $N_m \sim \text{Uniform}_D(1,100)$.
We estimate calibration on $5,000$ held-out simulations for each combination of $M$ and $N_m$.
As this implies simulating $50,000,000$ data sets, we forego the repetition procedure employed in the previous experiments.

\autoref{fig:variable_network_res} depicts the calibration and accuracy results for all combinations of $M$ and $N_m$.
We observe low ECEs for the vast majority of settings in \autoref{fig:eces_var_sizes} (median ECE over all settings of $\widehat{\text{ECE}} = 0.013$).
In other words, the trained network is capable of generating highly calibrated PMPs over a broad range of data set sizes.
Moreover, the BMC results are sensitive to the number of nested observations $N_m$, but not to the number of groups $M$, in our experimental setups.
The only systematic drop in calibration occurs for data sets containing just a few nested observations ($N_m \leq 5$).
Considering that we observed better calibration for this low number of observations in a network trained on data sets with varying $N_m$ (see \autoref{fig:eces_var_obs}), we surmise that the drop in the edge areas in \autoref{fig:eces_var_sizes} arises from the challenging learning task over vastly different data set sizes \autocite[a phenomenon known as \textit{amortization gap};][]{cremer2018inference}.  
The overall low (i.e., good) ECEs for all cases but the poorly identifiable $N_m = 1$ setting suggest that the networks' approximations are generally trustworthy.
This is further confirmed by \autoref{fig:accuracies_var_sizes}, where the observable accuracy pattern assures that this high calibration does not arise from a trade-off with predictive performance.
Despite the demanding amortization setting, the network achieves an excellent median accuracy of $\widehat{\text{Acc}} = .88$, similar to the earlier experiments. 
We also find no indication of bias in any of the test settings except the $N_m = 1$ setting (see \autoref{fig:SBC_variable_sizes}).
Marginal diagnostic plots for all metrics are provided in \autoref{fig:marginals_variable_sizes}.

\subsubsection{Bridge Sampling Comparison}
\label{sec:val1_bs_comp}

After validating the general trustworthiness of our method, we now benchmark it against the current gold standard for comparing HMs, namely, bridge sampling, as implemented by \cite{gronau2017Rbridgesampling}.
As the non-amortized nature of bridge sampling restricts the feasible number of test sets, we conduct the benchmarking on $100$ test sets which are simulated equally from \(\modelone\) and \(\modeltwo\).
All simulated data sets consist of $M = 50$ groups and $N_m = 50$ observations per group.
The fixed sample sizes of the test sets allow us to compare the two most distinct networks from \textbf{Section}~\ref{sec:val1_calibration} to bridge sampling: 
The \textit{fixed network} that is trained for this specific sample size and the more complex \textit{variable network} that is trained for amortized model comparison over variable sample sizes between $M = 1,\ldots,100$ groups and $N_m = 1,\ldots,100$ observations per group.

For bridge sampling, we first run four parallel MCMC chains with a warm-up period of $1,000$ draws and $49,000$ post-warm-up posterior draws per chain in Stan \autocite{carpenter2017stan, standev2019}.
We assess convergence through a visual inspection of the MCMC chains and an assessment of the $\widehat{R}$, bulk ESS and tail ESS metrics \autocite{vehtari2021rank}.
Afterwards, we use the posterior draws to approximate PMPs and BFs with the \textit{bridgesampling} R package \autocite{gronau2017Rbridgesampling}.
We confirm the sufficiency of the total of $196,000$ posterior draws by assessing the variability between multiple runs as in \textcite{schad2022workflow}, which yields highly similar results. 
Further insights via our calibration diagnostics are precluded by bridge sampling being a non-amortized method.

\paragraph{Approximation performance}
\label{sec:approx_performance}

\begin{table*}[h]
\begin{center}
\begin{minipage}{\textwidth}
\caption{Validation study 1: Performance metrics for the comparison between hierarchical normal models.}
\label{tab:comparison_metrics}
\begin{tabular*}{\textwidth}{@{\extracolsep{\fill}}lcccccc@{\extracolsep{\fill}}}
\toprule
 & Accuracy & MAE & RMSE & Log-Score & SBC \\
\midrule
Bridge sampling & 0.86 (0.03) & 0.19 (0.03) & 0.32 (0.03) & 0.32 (0.06) & -0.02 (0.04) \\
Fixed network & 0.84 (0.04) & 0.19 (0.03) & 0.32 (0.03) & 0.32 (0.06) & -0.01 (0.04) \\
Variable network & 0.86 (0.03) & 0.19 (0.03) & 0.32 (0.03) & 0.31 (0.06) & -0.01 (0.04) \\
\bottomrule
\end{tabular*}
\footnotetext{\textit{Note.} Bootstrapped mean values and standard errors (in parentheses) are presented. We use $1000$ bootstrap versions of the test data sets and estimate the standard errors from the bootstrap standard deviations of the metrics.}
\end{minipage}
\end{center}
\end{table*}

As we compare approximate PMPs, we can use a number of complementary metrics commonly employed to evaluate the quality of probabilistic predictions.
First, we quantify the fraction of times the correct model $\mathcal{M}^{(s)}_j$ underlying a simulated data set $s$ was detected, that is, the accuracy of recovery (see Equation~\ref{eq:accuracy}). 
Second, we assess the Mean Absolute Error (MAE) to investigate the average deviation of the approximated model probabilities $\widehat{\pi}^{(s)}_j$ from a perfect classification:
\begin{equation}
     \text{MAE}_j := {\frac{1}{S} \sum_{s=1}^{S} \left| \widehat{\pi}^{(s)}_j - \mathbb{I}_{\mathcal{M}_j}^{(s)} \right|}.
\end{equation}
Third, we measure the Root Mean Squared Error (RMSE), which places particular emphasis on large prediction errors, to detect whether one method produces highly incorrect approximations more frequently than the other:
\begin{equation}
     \text{RMSE}_j := \sqrt{\frac{1}{S} \sum_{s=1}^{S} \left( \widehat{\pi}^{(s)}_j - \mathbb{I}_{\mathcal{M}_j}^{(s)} \right)^2}.
\end{equation}
Fourth, we calculate the Log-Score following the logarithmic scoring rule:
\begin{equation}
    \text{LogScore}_j := - \frac{1}{S} \sum_{s=1}^{S} \left[ \mathbb{I}_{\mathcal{M}_j}^{(s)} \cdot  \text{log} \widehat{\pi}^{(s)}_j \right].
\end{equation}
Its property as a strictly proper scoring rule implies that it is asymptotically minimized if and only if the approximate probabilities equal the true probabilities \autocite{gneiting2007strictly}. 
Lastly, we measure simulation-based calibration \autocite[SBC;][]{talts2018validating} as adapted by \textcite{schad2022workflow} for model inference by the difference between the prior probability for a model and its average posterior probability in the test sets:
\begin{equation}
    \text{SBC}_j := p(\model) - \frac{1}{S} \sum_{s=1}^{S} \widehat{\pi}^{(s)}_j.
\end{equation}
We evaluate all metrics for \(\modeltwo\), so that a bias towards \(\modelone\) is indicated by positive SBC values and a bias towards \(\modeltwo\) by negative SBC values.

Table~\ref{tab:comparison_metrics} depicts the comparison results for our experimental setting.
All metrics show equal performances for bridge sampling and the two neural network variants, with any differences being well within the range of the standard errors. 

\paragraph{Approximation convergence}

\begin{figure*}
    \centering
    \includegraphics[width=0.95\textwidth]{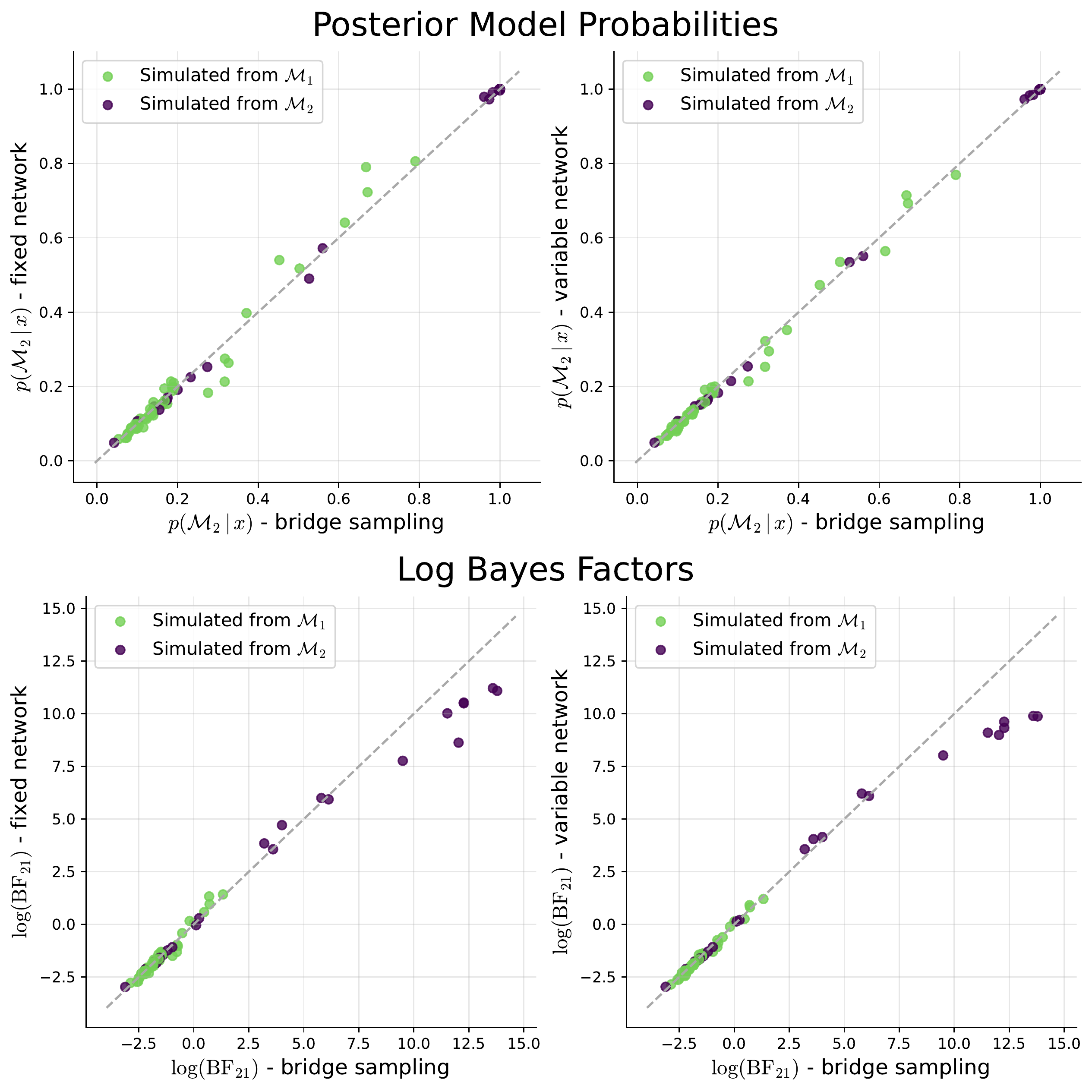}
    \caption{Validation study 1: Comparison of approximation results obtained via bridge sampling vs. the neural network trained on fixed data set sizes (left) and the neural network trained on variable data set sizes (right). For visibility purposes, the Bayes factor plots include only those 73 data sets for which bridge sampling approximated a $\mathrm{BF}_{21} <1,000,000$ (plots with all data sets are provided in Appendix~\ref{app:val1_details}).}
    \label{fig:BS_vs_NN_convergence}
\end{figure*}

In the following, we analyze the degree of convergence between the two methods at the level of individual data sets.
We explore this visually by contrasting the PMP and (natural logarithmic) BF approximations of bridge sampling with the two neural network variants in \autoref{fig:BS_vs_NN_convergence}.
We observe that the two methods' PMP approximations agree for the easy cases where the true underlying model is clearly classifiable.
Thus, discrepancies between the two methods arise mainly for data sets with predicted PMPs close to $\widehat{\pi}=0.5$. 
Even for the data sets with the largest discrepancies, the two methods do not map to qualitatively different decisions: 
$\widehat{\pi}^{(\text{bridge})}_2 = .67$ and  $\widehat{\pi}^{(\text{neural})}_2 = .79$ for the fixed network, $\widehat{\pi}^{(\text{bridge})}_2 = .32$ and  $\widehat{\pi}^{(\text{neural})}_2 = .25$ for the variable network.
Most importantly, we detect no systematic pattern in these deviations.

As BFs represent the ratio of marginal likelihoods, they allow for a closer inspection of the degree of agreement between the methods in those edge cases with PMPs close to $0$ or $1$.
We observe a close convergence for data sets classified as stemming from \(\modelone\).
Considering the predictions favoring \(\modeltwo\), there are discrepancies for data sets with log BFs $> 9.49$.
Since this corresponds to BFs $> 13,000$ and PMPs $> .9999$, it is not visible in the PMP approximation plots.
We obtain such extreme results only for \(\modeltwo\), as this model allows for deviations of the group level parameters' location from $0$ and enables the occurrence of extreme evidence in its favor.
The divergence in this area of extreme evidence emerges most likely from the loss function employed for training the neural networks: 
The logarithmic loss obtained from a minuscule deviation of the PMP from $1$ is near $0$, which results in a negligible incentive for further optimization of the network's weights. 
We could reject a competing explanation based on limited floating-point precision, since training with an increased floating-point precision from $32$-bit to $64$-bit resulted in identical patterns.
For visibility purposes, we exclude the 27 data sets for which bridge sampling approximated a BF $>1,000,000$ for the BF plots in \autoref{fig:BS_vs_NN_convergence}, all continuing the observed plateau pattern. 
Plots with all 100 data sets are provided in Appendix~\ref{app:val1_details}.

The divergence we encountered provides insights into the technical nature of our method but only arises in cases of extreme evidence. 
Thus, it is far from altering the substantive conclusions derived from the simulated BMC setting.
Considering the convergence between the two methods in the realm of practical relevance, we can conclude that our method produces highly similar approximations to bridge sampling in this scenario.

\paragraph{Approximation time}
\label{sec:val1_approx_times}

Both bridge sampling and our deep learning method can be divided into two computational phases.
For bridge sampling, the first phase consists of drawing from the posterior parameter distributions (taking $52$ seconds per data set on average).
Bridge sampling itself takes place in the second phase (taking $38$ seconds on average). 
Notably, in contrast to amortized inference with neural networks, both phases need to be repeated for each (simulated or observed) data set.
Taking the initial compilation time of $42$ seconds into account, bridge sampling consequently took $152$ minutes for BMC on our $100$ test data sets.

For the neural networks, the first phase (training) is resource-intensive (taking $6$ minutes for the fixed network and $36$ minutes for the variable network).
The second phase (inference) is then performed in near real-time (inference on all $100$ test data sets took $0.0004$ seconds for the fixed network and $0.007$ seconds for the variable network) and thus amortizes the training cost over multiple applications. 
For the simple HMs compared here, the amortization gains of our networks over bridge sampling come into effect after performing BMC on $4$ (fixed network) or $24$ (variable network) data sets.

We acknowledge our likely suboptimal choices of computational steps for the bridge sampling workflow or the neural networks and hence wish to stress the general patterns of non-amortized vs. amortized methods demonstrated here. 
In general, we expect an advantage of bridge sampling in terms of efficiency in situations where only one or a few data sets are available and obtaining a large number of posterior draws is feasible.
The demonstrated amortization property of our method might not be so relevant for inference on a single hierarchical data set, but it becomes crucial for performing calibration or recovery studies, which necessitate multiple re-fits of the same model \autocite{schad2022workflow}.

\begin{figure*}
    \centering
    \includegraphics[width=\textwidth]{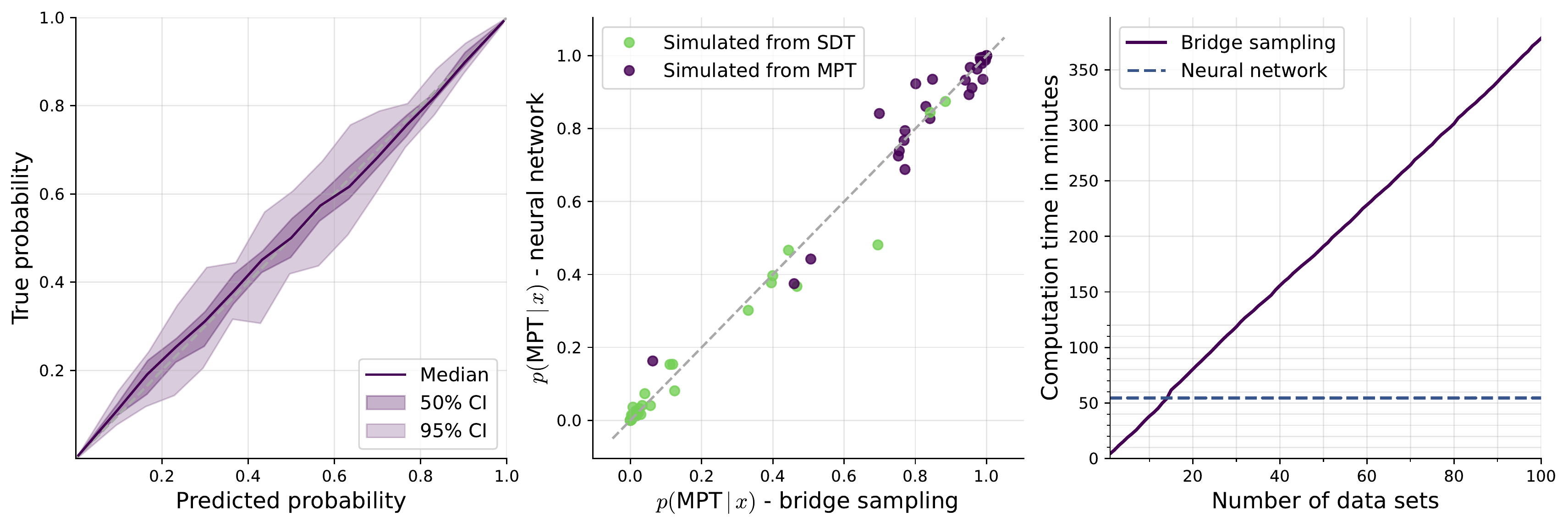} \\[-.02\linewidth]
    \subfloat[\label{fig:calibration_SDTMPT} Calibration of the neural network over $25$ repetitions with $5,000$ data sets each.]{\hspace{.34\linewidth}} \hspace{.01\linewidth}
    \subfloat[\label{fig:BS_vs_NN_SDTMPT_PMPs} Convergence of approximate PMPs.]{\hspace{.31\linewidth}} \hspace{.01\linewidth}
    \subfloat[\label{fig:BS_vs_NN_SDTMPT_comp_times} Computation times.]{\hspace{.32\linewidth}}
    \caption{Validation study 2: Results for the comparison between hierarchical SDT and MPT models.}
    \label{fig:BS_vs_NN_SDTMPT}
\end{figure*}

\begin{table*}[h]
\begin{center}
\begin{minipage}{\textwidth}
\caption{Validation study 2: Performance metrics for the comparison between hierarchical SDT and MPT models.}
\label{tab:comparison_metrics_SDTMPT}
\begin{tabular*}{\textwidth}{@{\extracolsep{\fill}}lcccccc@{\extracolsep{\fill}}}
\toprule
 & Accuracy & MAE & RMSE & Log Score & SBC \\
\midrule
Bridge sampling & 0.95 (0.02) & 0.1 (0.02) & 0.22 (0.03) & 0.16 (0.04) & -0.01 (0.04) \\
Neural network & 0.95 (0.02) & 0.1 (0.02) & 0.21 (0.03) & 0.16 (0.04) & 0.00 (0.04) \\
\bottomrule
\end{tabular*}
\footnotetext{\textit{Note.} Bootstrapped mean values and standard errors (in parentheses) are presented. We use $1000$ bootstrap versions of the test data sets and estimate the standard errors from the bootstrap standard deviations of the metrics.}
\end{minipage}
\end{center}
\end{table*}

\subsection{Validation Study 2: Hierarchical SDT vs. MPT Models}
\label{subsec:val2}
We now extend our validation experiments from the simple setup with nested HMs to the comparison of non-nested HMs of cognition. 
In this simulation study, we examine the ability of our method to distinguish between data sets generated either from an HM based on signal detection theory \autocite[SDT model;][]{green1966signal} or a hierarchical multinomial processing tree model \autocite[MPT model;][]{riefer1988multinomial}.
For illustrative purposes, we embed our simulation study within an old-new recognition scenario, where participants indicate whether or not a stimulus was previously presented to them.

We ensure a challenging model comparison setting via three design aspects:
First, we specify both models to possess a similar generative behavior, that is, hardly distinguishable prior predictive distributions of hit rates and false alarm rates (prior predictive plots are provided in Appendix~\ref{sec:val2_details}).
Second, data sets of old-new recognition typically contain low information as they only consist of binary variables indicating the stimulus type and response, respectively. 
Third, we further amplify the information sparsity of the data sets by choosing a particularly small size for all data sets of $M = 25$ simulated participants and $N_m = 50$ observations per participant.

A major difference between the compared cognitive model classes lies in the assumption of a continuous latent process by the SDT model and discrete processes (or states) by the MPT model.
Our specification of the SDT model follows the hierarchical formulation of the standard equal-variance model by \textcite{rouder2005introduction}.
As the competing MPT model, we specify a hierarchical latent-trait two-high-threshold model \autocite{klauer2010hierarchical}, which, in contrast to the SDT model, explicitly models correlations between its parameters. 
We follow the convention of restricting the parameters that describe the probability of recognizing a previously presented stimulus as old and a distractor stimulus as new to be equal, $D_{O} = D_{N}$, to render the MPT model identifiable \autocite{singmann2013mptinr, erdfelder2009multinomial}.
Our prior choices for the parameters of both models are described in Appendix~\ref{sec:val2_details}.

We train the neural network for $50,000$ training steps. 
As in \textbf{Section}~\ref{subsec:val1}, we first leverage the amortization property of our method to inspect its calibration for the current model comparison task.
\autoref{fig:calibration_SDTMPT} shows that the trained neural network generates well-calibrated PMP approximations (median ECE over $25$ repetitions of $\widehat{\text{ECE}} = 0.009$).

Next, we assess whether the observed calibration of the network translates into a competitive performance relative to bridge sampling.
The benchmarking setup ($50$ simulated data sets from each model) and the implementation of the bridge sampling workflow follow the procedure described in \textbf{Section}~\ref{sec:approx_performance}.

The classification metrics depicted in Table~\ref{tab:comparison_metrics_SDTMPT} reveal the excellent performance of both methods, despite the challenging BMC scenario. 
We further observe a high degree of convergence between approximate PMPs derived by the two methods (cf.~\autoref{fig:BS_vs_NN_SDTMPT_PMPs}).
Again, we find discrepancies between bridge sampling and our method in areas of extreme evidence (see \autoref{fig:BS_vs_NN_SDTMPT_BFs_full} for log BFs).
As depicted in \autoref{fig:BS_vs_NN_SDTMPT_comp_times}, obtaining PMP approximations for the $100$ test data sets took more than $6$ hours for bridge sampling and $55$ minutes for the neural network.
For this comparison of more complex cognitive models than in \textbf{Section}~\ref{sec:val1_bs_comp}, the amortization advantage of our method emerges when analyzing $15$ or more data sets. 
Note that this advantage would quickly show up in validation studies involving multiple model re-fits (e.g., bootstrap, sensitivity analysis or cross-validation).

All validation experiments so far have been set up in an $\mathcal{M}$-closed setting, with validation data simulated from the set $\mathcal{M}$ of models under consideration \autocite{bernardo1994bayesian}.
Therefore, as a final validation, we test whether our method also behaves sensibly in an $\mathcal{M}$-open setting, where none of the models generated the test data.
For this, we simulate $100$ noise data sets with the same hierarchical structure as before but generate the binary values for stimulus types and responses from a Bernoulli distribution with $p = 0.5$.
Our neural method agrees with bridge sampling by assigning very high PMPs to the SDT model for all noise data sets ($\bar{\pi}^{(\text{bridge})}_{SDT} = .999965$; $\bar{\pi}^{(\text{neural})}_{SDT} = .999958$).
Correspondingly, the deviations between both methods are minimal.
We thus observe a close alignment between bridge sampling and our neural method in both a well-specified and a misspecified scenario.
This tentative result suggests that our amortized estimates are faithful approximations not only in an $\mathcal{M}$-closed but also an $\mathcal{M}$-open setting, at least for this BMC scenario.

The converging results from the two validation studies demonstrate that our neural method generates well-calibrated and accurate PMP approximations.
Despite our method only accessing the likelihood function indirectly via simulations, it can successfully compete with bridge sampling, which has direct access to the likelihood function. 

\subsection{Application: Hierarchical Evidence Accumulation Models}
\label{subsec:application}

In the following, we showcase the utility of our method by comparing complex hierarchical EAMs in a real-data situation where likelihood-based methods such as bridge sampling would not be applicable. 
More precisely, we seek to test the explanatory power of different stochastic diffusion model formulations proposed by \textcite{voss2019sequential} for experimental response time data.

The so-called Lévy flight model increases the flexibility of the standard Wiener diffusion model \autocite{ratcliff2016diffusion} but renders its likelihood function intractable with standard numerical approximations \autocite{voss2007fast}.
The complete incorporation of all information through hierarchical modeling and the realization of BMC has consequently been infeasible so far.
Thus, in a recent study, \textcite{wieschen2020jumping} had to resort to a separate computation of the Bayesian Information Criterion (BIC) for each participant with subsequent aggregation.
We aim to extend the study of \textcite{wieschen2020jumping} by comparing fully hierarchical EAMs through PMPs and BFs.
Moreover, we intend to answer the question formulated by \cite{wieschen2020jumping} as to whether the superior performance of the more complex models in their study stems from an insufficient punishment of model flexibility by the BIC.
In addition to addressing a substantive research question in this application, we also demonstrate multiple advantages of our deep learning method on empirical data:

\begin{itemize}
  \item \textit{Compare HMs with intractable likelihoods}: As our method is simulation-based, including models with intractable likelihood functions in the comparison set does not alter its feasibility.
  \item \textit{Adequately model nested data}: Our method alleviates computational challenges that prevent modelers from adequately capturing the information contained in nested data structures through HMs.
  \item \textit{Re-use trained networks via fine-tuning}: We accelerate the training of our neural network by pre-training it on less complex simulated data and subsequently fine-tuning it on simulated data resembling the actual experimental setting.
  \item \textit{Handle missing data}: We train a neural network that can handle varying amounts of missing data by randomly masking simulated data during the training process.
  \item \textit{Validate a trained network on simulated data}: The amortized nature of our method allows for extensive validation of a trained network prior to its application to empirical data.
\end{itemize}

\begin{figure*}
    \centering
    \includegraphics[width=\textwidth]{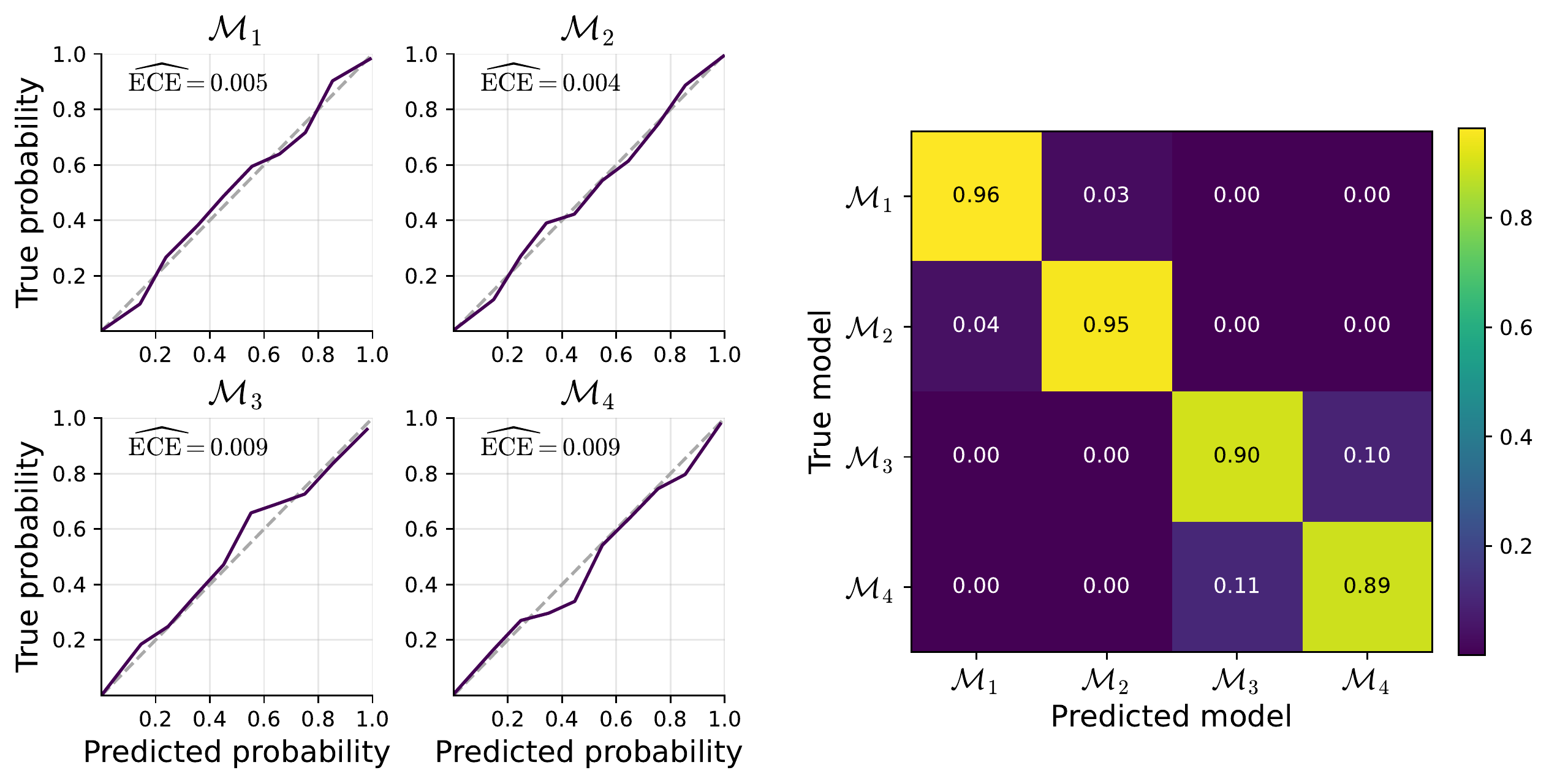} \\[-.02\linewidth]
    \subfloat[\label{fig:levy_validation_confusion} Calibration curves.]{\hspace{.5\linewidth}}
\subfloat[\label{fig:levy_validation_calibration} Confusion matrix.]{\hspace{.5\linewidth}}
    \caption{Real-data application: Validation results for the evidence accumulation models on $2,000$ simulated data sets per model.}
    \label{fig:levy_validation}
\end{figure*}

\subsubsection{Model Specification}

For this application, we consider a Lévy flight model with non-Gaussian noise \autocite{voss2019sequential}.
The Lévy flight process is driven by the following stochastic ordinary differential equation:
\begin{align}\label{eq:levy}
dx &= v\,dt + \sigma d\xi  \\
\xi &\sim \text{AlphaStable}(\alpha,\mu = 0,\sigma = \frac{1}{\sqrt{2}},\beta = 0), 
\end{align}
which represents a Lévy walk characterized by a fat-tailed stable noise distribution.
\footnote{An earlier version of this work used the original formulation by \textcite{voss2019sequential}, which sets $\sigma = 1$. For the special case of $\alpha = 2.0$, which is equivalent to the  Wiener diffusion model,  $\sigma = 1$ leads to an unusual diffusion constant (standard deviation of Gaussian noise) of $\sqrt{2}$, whereas $\sigma = \frac{1}{\sqrt{2}}$ ensures the conventional diffusion constant of $1$. Notably, model comparison results are highly sensitive to the choice of $\sigma$.}
In the above equation, $x$ denotes the accumulated (perceptual) evidence, $v$ denotes the rate of accumulation and $\alpha$ controls the tail exponent of the noise variate $\xi$.
\textcite{voss2019sequential} and \textcite{wieschen2020jumping} argue that the more abrupt changes in the information accumulation process that this model allows for could provide a better description of human decision-making than a Gaussian noise.
The addition of Lévy noise renders the standard numerical approximation of the diffusion model likelihood intractable \autocite{voss2007fast}.
Consequently, neither standard MCMC nor bridge sampling are applicable for Bayesian parameter estimation and BMC, respectively.

There is an ongoing debate about the inclusion of additional parameters that account for inter-trial variability in the diffusion model parameters: 
While they can provide a better model fit, the estimation of inter-trial variability parameters is often difficult and can result in unstable results  \autocite{boehm2018diffusion, lerche2016diffusion}.
Thus, \textcite{wieschen2020jumping} also compared basic (without inter-trial variability parameters) to full (with inter-trial variability parameters) versions of the drift-diffusion and Lévy flight model.

Consequently, the set of candidate models considered here consists of four EAMs with increasing flexibility (i.e., the scope of possible data patterns that they can generate):
\begin{itemize}
  \item \(\modelone\), the most parsimonious \textit{basic diffusion model} with the parameter $v$ describing the mean rate of information uptake, the parameter $a$ describing the threshold at which a decision is made, the parameter $z_r$ describing a bias of the starting point towards one decision alternative and the parameter $t_0$ describing the non-decision time, that is, the time spent encoding the stimulus and executing the decision.
  \item \(\modeltwo\), the \textit{basic Lévy flight model}, in which the assumption of a Wiener diffusion process with Gaussian noise is replaced by the above introduced Lévy flight process. 
  The additional free parameter $\alpha$ governs the tail behavior of the noise distribution. 
  The setting $\alpha = 2$ is equivalent to a Gaussian distribution, whereas $\alpha = 1$ reduces to a Cauchy distribution.
  \item \(\modelthree\), the \textit{full diffusion model}, which extends \(\modelone\) with the parameters $s_{v_m}$, $s_{z_m}$ and $s_{t_m}$ that denote the variability (i.e., standard deviations) of drift rate, starting point bias and non-decision time, respectively, between trials.
  \item \(\modelfour\), the \textit{full Lévy flight model}, that possesses the largest flexibility by including inter-trial variability parameters as well as the flexible Lévy noise distribution controlled by $\alpha$.
\end{itemize}

\noindent Parameter priors and prior predictive checks are provided in Appendix~\ref{subsec:app_priors}.

\subsubsection{Data}

The reanalyzed data set by \textcite{wieschen2020jumping} contains $40$ participants who completed a total of $900$ trials of binary decision tasks (color discrimination and lexical decision) each. 
On average, $3.17\%$ of trials per participant were excluded due to extremely short or long reaction times.

\begin{figure*}[h]
    \centering
    \includegraphics[width=\textwidth]{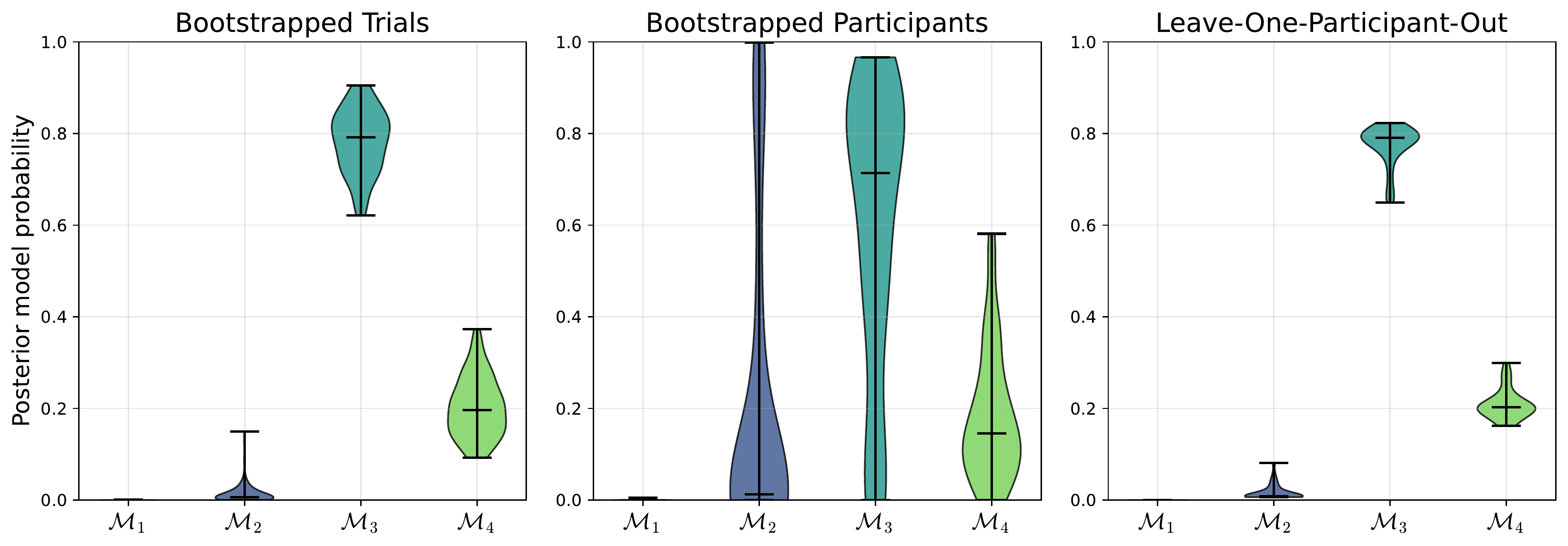}
    \caption{Real-data application: Model posteriors on the empirical data set with uncertainty under different data perturbations. We use $100$ bootstrap samples for the bootstrapped results.}
    \label{fig:diffusion_posterior}
\end{figure*}

\subsubsection{Simulation-Based Training} 
\label{subsubsec:application_training}

Since simulating data from EAMs can be challenging, especially when they include non-Gaussian noise, we leverage the advantage that neural networks are capable of transfer learning as described in \textbf{Section}~\ref{subsec:learning_BMC}.
Transfer learning describes the utilization of representations that had been previously learned by a neural network in a particular task for a new, related task \autocite[e.g.,][]{ng2015deep}.
In this way, neural networks can be applied in small data settings (e.g., a limited simulation budget) by re-using the training knowledge encoded from structurally similar (possibly big data) settings.

For the purpose of model comparison, we first pre-train the network for $20$ epochs (passes over the whole training data) on $10,000$ simulated data sets per model.
These data sets resemble the empirical data in that they consist of $40$ simulated participants, but differ in that the number of trials is reduced by a factor of 9 ($100$ instead of $900$ trials per participant).
Afterwards, we fine-tune the network for additional $30$ epochs on $2,000$ simulated data sets per model that match the empirical data set with $40$ simulated participants and $900$ trials per participant.
Thereby, we considerably reduce the computational demand of the training process.
We further speed up the training phase by simulating all data prior to the training of the network in the high-performance programming language Julia \autocite{bezanson2017julia}.
Pre-training took 10 minutes for the simulations and 11 minutes for training the networks.
Fine-tuning took 18 minutes for the simulations and 16 minutes for training the networks, resulting in a total of 55 minutes for the training phase.

To fully adapt the network to the characteristics of the empirical data, we also simulate missing data during fine-tuning.
In each training epoch, we generate a random binary mask $\bs{f}$ coding the simulated missing values.
We sample the number of masked trials from a (discretized) normal distribution truncated between $1$ and the number of trials, $900$.
The distributions' mean and standard deviation match the amount and variability of missing trials in the empirical data.
We then perform an element-wise multiplication $\tilde{\xb} = \xb \otimes \bs{f}$ and feed the ``contaminated'' data $\tilde{\xb}$ to the network.
This procedure results in a robust network that can process various proportions of missing data.
We find rank stability of our results in the presence of up to $25\%$ missing data in Appendix~\ref{subsec:app_robustness}.
 
\subsubsection{Results}

Before applying our trained network to the empirical data, we validate it on $2,000$ simulated data sets per model. 
First, the individual calibration curves in \autoref{fig:levy_validation_confusion} show excellent calibration for all models with $\widehat{\text{ECEs}}$ close to $0$.
The calibration curves now consist of $10$ instead of $15$ intervals to obtain stable results despite the smaller amount of validation data sets per model.
Second, we evaluate the accuracy of recovery and patterns of misclassification through the confusion matrix depicted in \autoref{fig:levy_validation_calibration}.
The confusion matrix confirms that the excellent calibration of the network does not stem from chance performance.
It also reveals that the selection of the ``true'' model becomes more difficult with increasing model complexity, which is a direct consequence of the Occam's razor property inherent in BMC (cf.~\autoref{fig:bmc}).

\begin{table}[h]
\centering
\caption{Real-data application: Bayes factors (BFs) and posterior model probabilities (PMPs) estimated from data by \textcite{wieschen2020jumping}. The preferred model is indicated by an asterisk.}
\label{tab:diffusion_results}
\begin{tabular}{lcccc}
\toprule
 & $\mathcal{M}_1$ & $\mathcal{M}_2$ & $\mathcal{M}_3$ & $\mathcal{M}_4$ \\
\midrule
$\text{BF}_{j3}$ & 9.63e-05 & 0.01 & * & 0.27 \\
$\text{BF}_{3j}$ & 1.04e+04 & 78 & * & 3.71 \\
$\text{PMP}$ & 7.51e-05 & 1.00e-02 & 0.78 & 0.21 \\
\bottomrule
\end{tabular}
\end{table}

Table~\ref{tab:diffusion_results} presents the model comparison results on the empirical data set.
Additionally, \autoref{fig:diffusion_posterior} displays the model posteriors under different data perturbations.
Consistent with the results of the non-hierarchical BIC approach by \textcite{wieschen2020jumping}, we find little evidence for both the basic diffusion model \(\modelone\) and the basic Lévy flight model \(\modeltwo\).
This implies that the additional complexity of allowing parameters to vary between trials in \(\modelthree\) and \(\modelfour\) is, even under the strict penalization of prior-predictive flexibility in BMC, outweighed by better model fit.
Also in agreement with \textcite{wieschen2020jumping}, we observe evidence for both \(\modelthree\) and \(\modelfour\), but, in contrast to \textcite{wieschen2020jumping}, our results slightly favor the full diffusion model \(\modelthree\) over the full Lévy flight model \(\modelfour\).
\autoref{fig:diffusion_posterior} confirms both the slight advantage of \(\modelthree\) over \(\modelfour\) and the substantial uncertainty associated with these results.

\section{Discussion} 
\label{sec:discussion}

Nested data are ubiquitous in the quantitative sciences, including psychological and cognitive research \autocite{farrell2018computational}. 
Yet, to avoid dealing with the complex dependencies resulting from these data, researchers often resort to simpler analyses, ignoring potentially important structural information. 
Hierarchical models (HMs) provide a flexible way to represent the multilevel structure of nested data, but this flexibility can make Bayesian model comparison a daunting undertaking. 

In this work, we proposed a powerful remedy to this problem: Building on the BayesFlow framework \autocite{radev2020bayesflow}, we developed a neural network architecture that enables approximate BMC for arbitrarily complex HMs. 
In two simulation studies, we showed that our deep learning method is well-calibrated and performs as accurately as bridge sampling, which is the current state-of-the-art for comparing HMs with simple likelihoods.
Moreover, in a subsequent real-data application, we compared the relatively new Lévy flight model with existing evidence accumulation models.
Thus, we argue that our method is well-suited to enhance the applicability of (complex) HMs in psychological research. 
Below, we summarize the key properties and limitations of our method while also outlining future research directions. 


\subsection{Amortized Inference}

Our method offloads the computational demands for comparing HMs onto the training phase of a custom neural network, allowing for near real-time model comparison using the trained network.
The resulting \textit{amortization} offers several advantages over non-amortized methods.

First, it enables thorough validation of a trained network on thousands of simulated data sets, allowing large-scale simulation-based diagnostics to become an integral part of the BMC workflow \autocite{schad2022workflow, gelman2020bayesian}.
Second, the trained and validated network can be used not only for point estimates of BFs or PMPs on empirical data but also for exploring the robustness of the results against multiple data perturbations, as showcased in our real-data application.

Third, we demonstrated the feasibility of amortizing over variable data set sizes in our first validation study.
This is particularly advantageous in the context of HMs since nested data sets often contain multiple exchangeable levels with variable sizes (e.g., different numbers of clusters, participants and observations).
Analyzing multiple hierarchical data sets with variable sizes only requires a single network that has seen different data set sizes during training.
The same network could also be used for various simulation studies, such as the challenging task of designing maximally informative experiments in a hierarchical BMC setting \autocite{myung2009optimal, heck2019maximizing}.

Lastly, we showed that researchers do not even need to consider all possible shapes of future data sets when training such a network, as they can use transfer learning to efficiently adapt a trained network to a related setting.
Beyond allowing more flexibility in reusing networks across experiments, researchers or even fields, transfer learning can also considerably reduce the computational demands associated with comparing complex HMs.
As demonstrated in our real-data application, a network can be pre-trained on simulated data sets with reduced size and fine-tuned afterwards on sizes matching the empirical data.


\subsection{Independence From Explicit Likelihoods}

Unlike other popular methods for performing BMC on HMs, such as the Savage-Dickey density ratio or bridge sampling, our method is not constrained by the availability of an explicit likelihood function for all competing models.
As long as the models in question can be implemented as simulators, the neural network can be trained to perform BMC on these models.
The value of such a method is evident, as it decouples the substantive task of model specification from concerns about the feasibility of estimation methods.

Statistical models are instantiations of substantive knowledge or hypotheses. 
As such, we argue that model specification should not be unduly restricted by considerations of computational tractability -- a sentiment that is closely related to what \textcite{haaf2021bayes} call the ``specification-first-principle''. 
Our proposed deep learning method satisfies this principle, as model specification may be guided exclusively by substantive arguments with few concerns about tractability.
Thus, we believe that our method makes a contribution to the recent upsurge of innovative psychological models \autocite{ghaderi2023general, heathcote2022winner, collins2022advances} by allowing for an efficient assessment of their incremental value in a hierarchical setting.

\subsection{Limitations and Outlook}

One of the main challenges of approximate methods and, more broadly, statistical inference is ensuring the faithfulness of the obtained results.
The outlined possibilities for validating the network and examining the robustness of the results are important contributions of our method but come with open questions.
Concerning the validation of the network, framing model comparison as a supervised learning problem allows us to draw from the rich literature on classification performance metrics.
Nevertheless, determining a ``good-enough'' score for an approximate BMC method remains challenging, as the optimally possible performance is application-specific and usually unknown.

Concerning the application of the network to empirical data, we showed in Validation Study 2 that our method produces, at least in this scenario, reasonable results when confronted with data not stemming from the models under consideration. 
Moreover, our robustness checks are a practical proxy for measuring the reliability of BMC results in a closed-world setting.
However, these checks cannot possibly capture the (lack of) absolute evidence for an HM:
As a relative method, BMC may indicate that one model fits the data \emph{better} than a set of competing models, but it does not provide any measure of how well (or poorly) the model itself approximates the underlying data-generating process.
A promising direction to address this limitation could be the combination of our method with the recently proposed meta-uncertainty framework for BMC \autocite{schmitt2022meta}, which can be greatly accelerated with amortized deep learning methods.
This combination could provide a principled delineation of different uncertainty sources, enabling the detection of \emph{model misspecification} cases where none of the competing HMs can explain the observed data.
Still, further research is needed to determine whether meta-uncertainty can provide reliable evidence for the open vs. closed world assumption in the context of HMs and prevent the dangers that simulation gaps (i.e., as induced by model misspecification) pose for simulation-based inference \autocite{hermans2021averting}.

Since BMC is a marginal likelihood (i.e., prior predictive) approach, the priors should be informed by scientific theory and will thus have a decisive influence on the results \autocite{vanpaemel2010prior}.
We do not intend to re-iterate the ongoing discussion about this property of BMC \autocite{haaf2021bayes, vehtari2019limitations, gronau2019limitations, gronau2019rejoinder}, but want to highlight a specific difficulty that arises for HMs:
Parameter priors of an HM are connected via multilevel dependencies, increasing the risk that poor prior choices lead to non-intended model behavior \autocite[for a recent discussion of this problem in cognitive modeling, see][]{sarafoglou2022theory}.
Therefore, prior predictive checks and prior sensitivity analyses become especially important when conducting BMC on competing HMs.
While transfer learning reduces the computational demands of retraining a neural network for sensitivity analyses, another avenue for future research would be the amortization over different prior choices, enabling immediate prior sensitivity assessment.

Finally, it should be noted that the version of our method explored here can only compare HMs assuming exchangeable data at each hierarchical level.
Although the majority of HMs in social science research follow this probabilistic symmetry, some researchers may want to compare non-exchangeable HMs, for example, to study within-person dynamics \autocite{schumacher2022neural, lodewyckx2011hierarchical, driver2018hierarchical}.
Fortunately, the modularity of our method allows easy adaptation of the neural network architecture to handle non-exchangeable HMs.
To compare hierarchical time series models with temporal dependencies at the lowest level, for instance, the first invariant module could be exchanged for a recurrent network, as proposed in \textcite{radev2021amortized}.
Thus, future research could extend and validate our method in BMC settings involving non-exchangeable HMs.



\section*{Acknowledgments}
Lasse Elsemüller was previously affiliated with the Department of Psychology, University of Mannheim, and Paul-Christian Bürkner with the Cluster of Excellence SimTech, University of Stuttgart.
The authors thank Lukas Schumacher for helpful comments on this manuscript.

LE and MS were supported by a grant from the Deutsche Forschungsgemeinschaft (DFG, German Research Foundation; GRK 2277) to the research training group Statistical Modeling in Psychology (SMiP).
LE was additionally supported by the Google Cloud Research Credits program with the award GCP19980904.
PCB was supported by the Deutsche Forschungsgemeinschaft under Germany’s Excellence Strategy – EXC-2075 - 390740016 (the Stuttgart Cluster of Excellence SimTech).
STR was supported by the Deutsche Forschungsgemeinschaft under Germany’s Excellence Strategy – EXC-2181 - 390900948 (the Heidelberg Cluster of Excellence STRUCTURES). 

\FloatBarrier
\printbibliography

\clearpage

\begin{appendices}
\section*{\centering{Appendix}}
\section{Neural Network Implementation and Training}
\label{sec:app_NN_training}

The neural networks are implemented in the Python library \textit{TensorFlow} \autocite{tensorflow2015} and jointly optimized via backpropagation.
During training, we use mini-batch gradient descent with batches of size $B = 32$ per backpropagation update (training step).
We employ the Adam optimizer \autocite{kingma2015adam} with a cosine decay schedule in Validation Study 1 (initial learning rate of $5 \times 10^{-4}$) and the real-data application (initial learning rate of $5 \times 10^{-4}$ for pre-training and $5 \times 10^{-5}$ for fine-tuning.
In Validation Study 2, we use the RMSprop optimizer \autocite{tieleman2012lecture} with an initial learning rate of $2.5 \times 10^{-4}$ and a cosine decay schedule, which we found to work better for the unusually sparse binary data.
For all validation studies, we use \textit{online training}, i.e. simulate new training data sets flexibly right before each training step.
For the real-data application, we simulate all data sets efficiently a priori in the Julia programming language and therefore use \textit{offline training}, i.e. training with a predetermined amount of data sets.

We use the following neural network architectures for all experiments: 
The hierarchical summary network is composed of two deep invariant modules, each consisting of $K=2$ equivariant modules followed by an invariant module. 
The inference network is realized via a standard feedforward network with three fully connected layers followed by a softmax output layer.
We did not conduct a thorough search for optimal hyperparameter settings of the neural networks and the training process.

\section{Validation Study 1 Details}
\label{app:val1_details}

\subsection{Calibration}
\label{app:val1_calibration}

\begin{figure*}[h]
    \centering
    \begin{subfigure}{0.48\textwidth}
     \includegraphics[width=\textwidth]{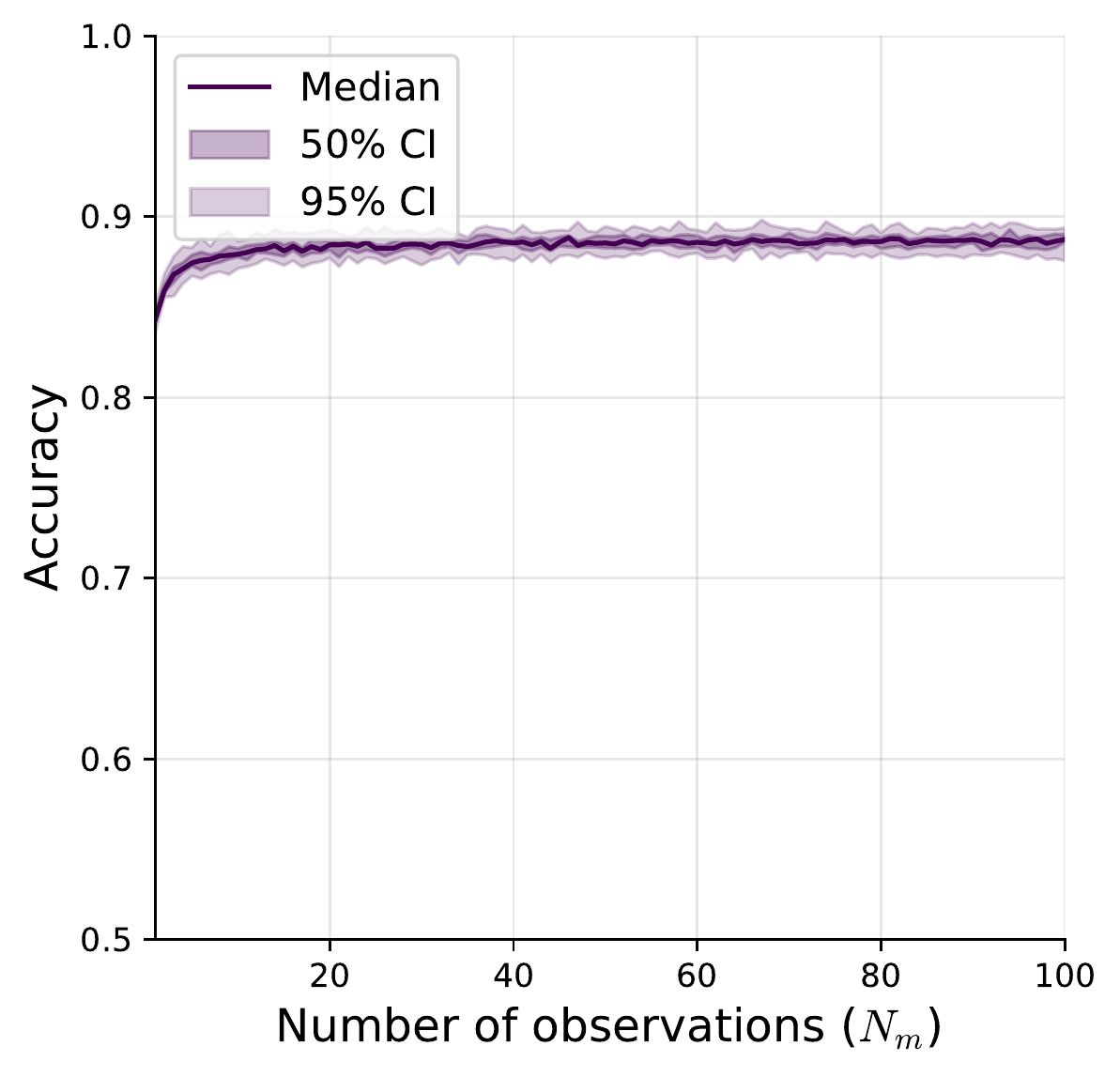}
     \subcaption{Accuracy of recovery.}
    \end{subfigure}
    \begin{subfigure}{0.51\textwidth}
     \includegraphics[width=\textwidth]{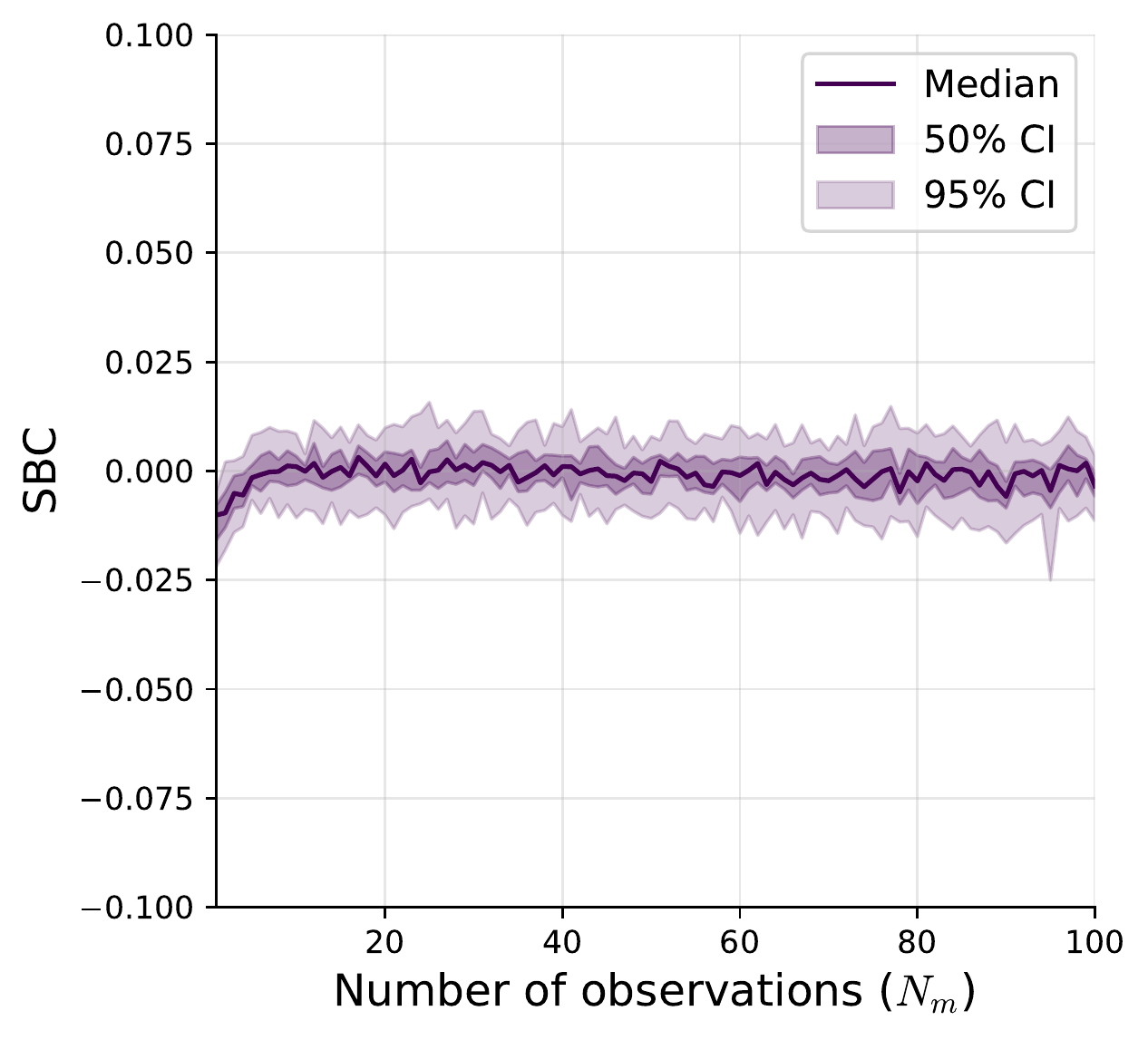}
      \subcaption{Simulation-based calibration (SBC).}
    \end{subfigure}
    \caption{Validation study 1: Additional results for the neural network trained and tested on data sets with varying numbers of observations.}    \label{fig:Acc_SBC_variable_observations}
\end{figure*}

Additional results for the scenario containing data sets with varying numbers of observations are depicted in \autoref{fig:Acc_SBC_variable_observations}. 
Accuracy and SBC (median of $\widehat{\text{SBC}} = -.0006$) are stable across nearly all settings, only slightly dropping for data sets with few observations.

\begin{figure*}[h]
    \centering
    \includegraphics[width=0.5\textwidth, trim=1cm 0cm 0cm 0cm]{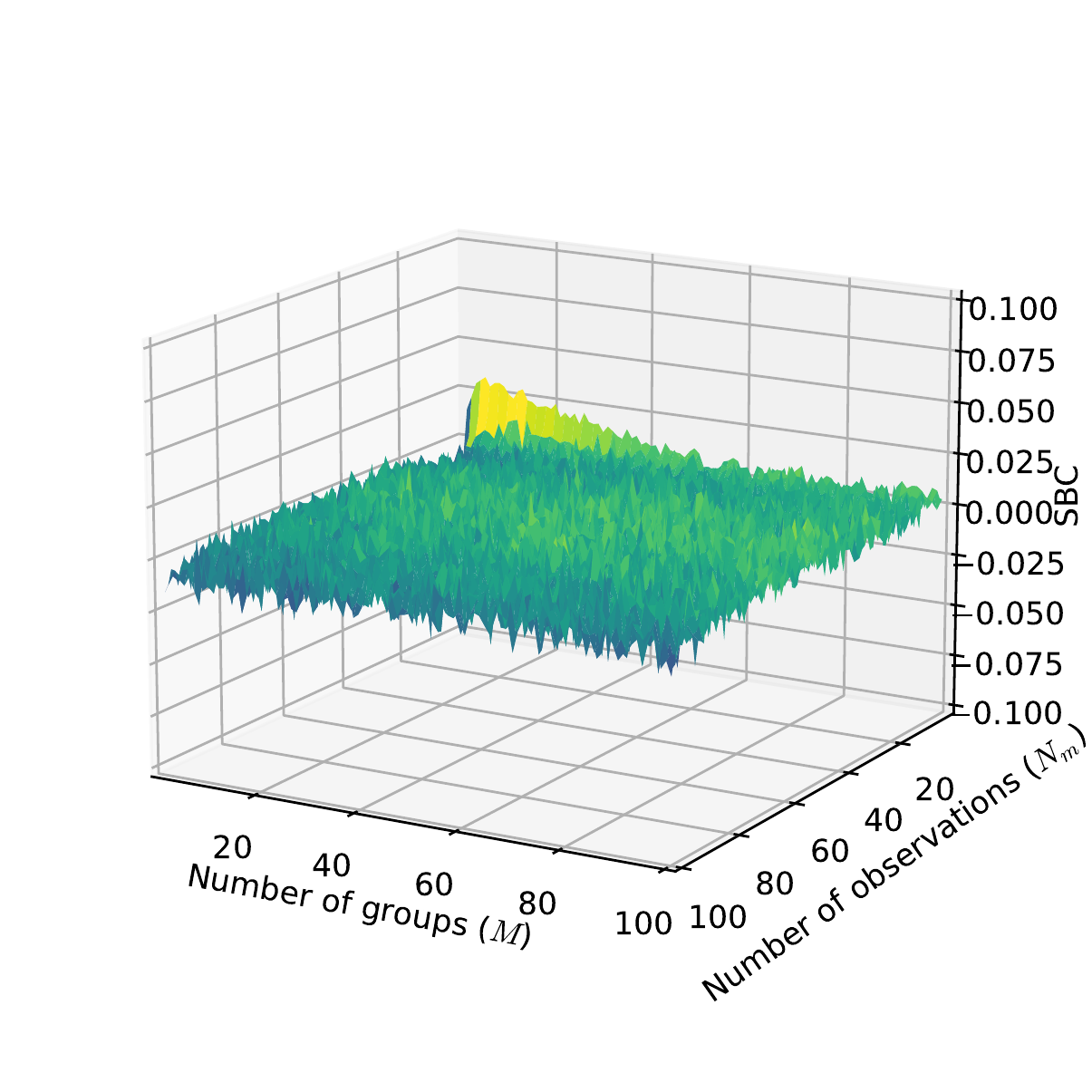}
    \caption{Validation study 1: SBC results for the neural network trained and tested over variable data set sizes.}
    \label{fig:SBC_variable_sizes}
\end{figure*}

\begin{figure*}[p]
    \centering
    \includegraphics[width=.85\textwidth]{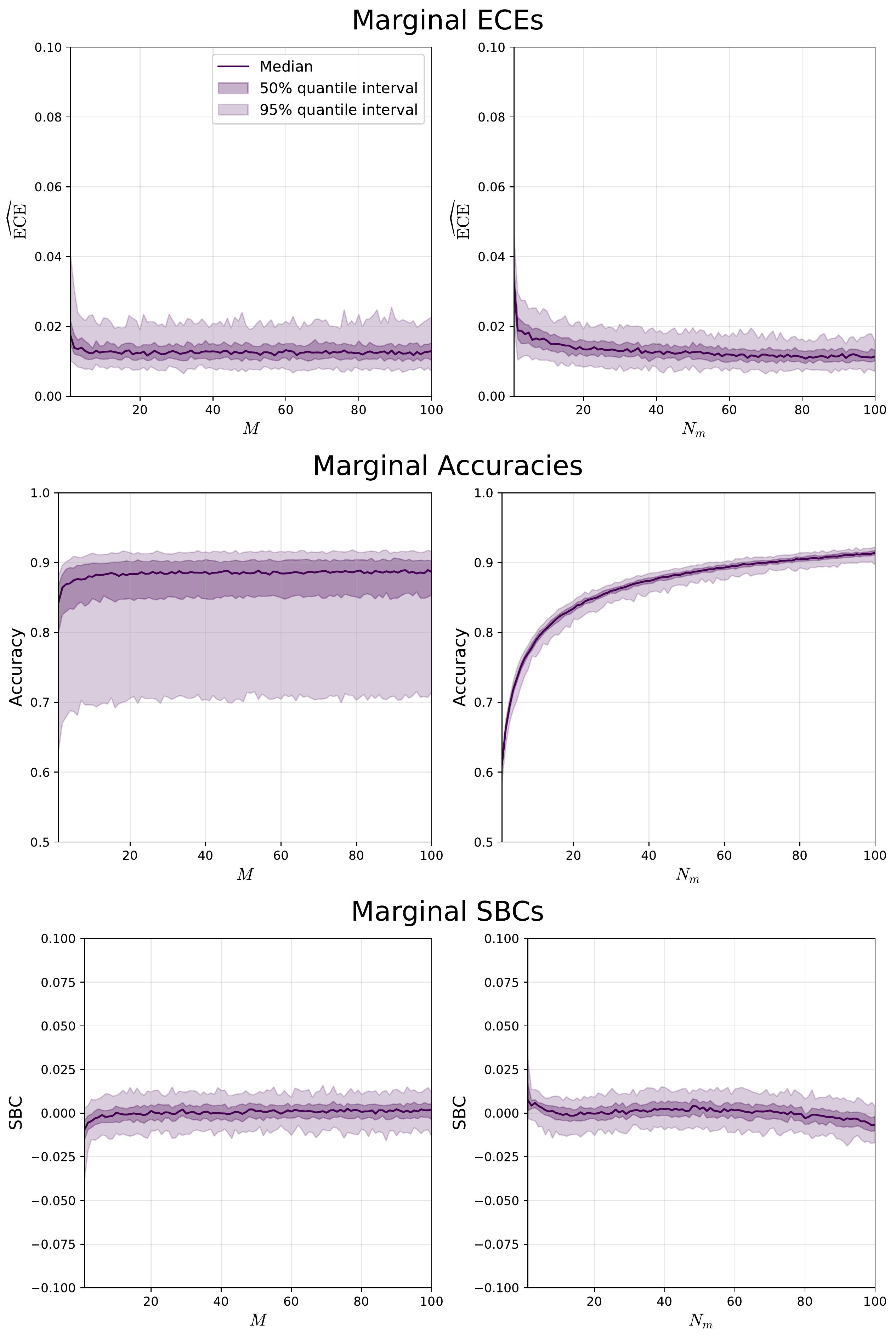}
    \caption{Validation study 1: Marginal plots for the neural network trained and tested over variable data set sizes.}
    \label{fig:marginals_variable_sizes}
\end{figure*}

Concerning the scenario containing data sets with varying numbers of groups and nested observations, \autoref{fig:SBC_variable_sizes} presents generally unbiased SBC results with a median of $\widehat{\text{SBC}} = .0004$.
\autoref{fig:marginals_variable_sizes} shows marginal plots corresponding to the 3D plots for all metrics.

\subsection{Bridge Sampling Comparison}
\label{app:val1_bs_comp}

\begin{figure*}[h]
    \centering
    \begin{subfigure}{0.49\textwidth}
     \includegraphics[width=\textwidth]{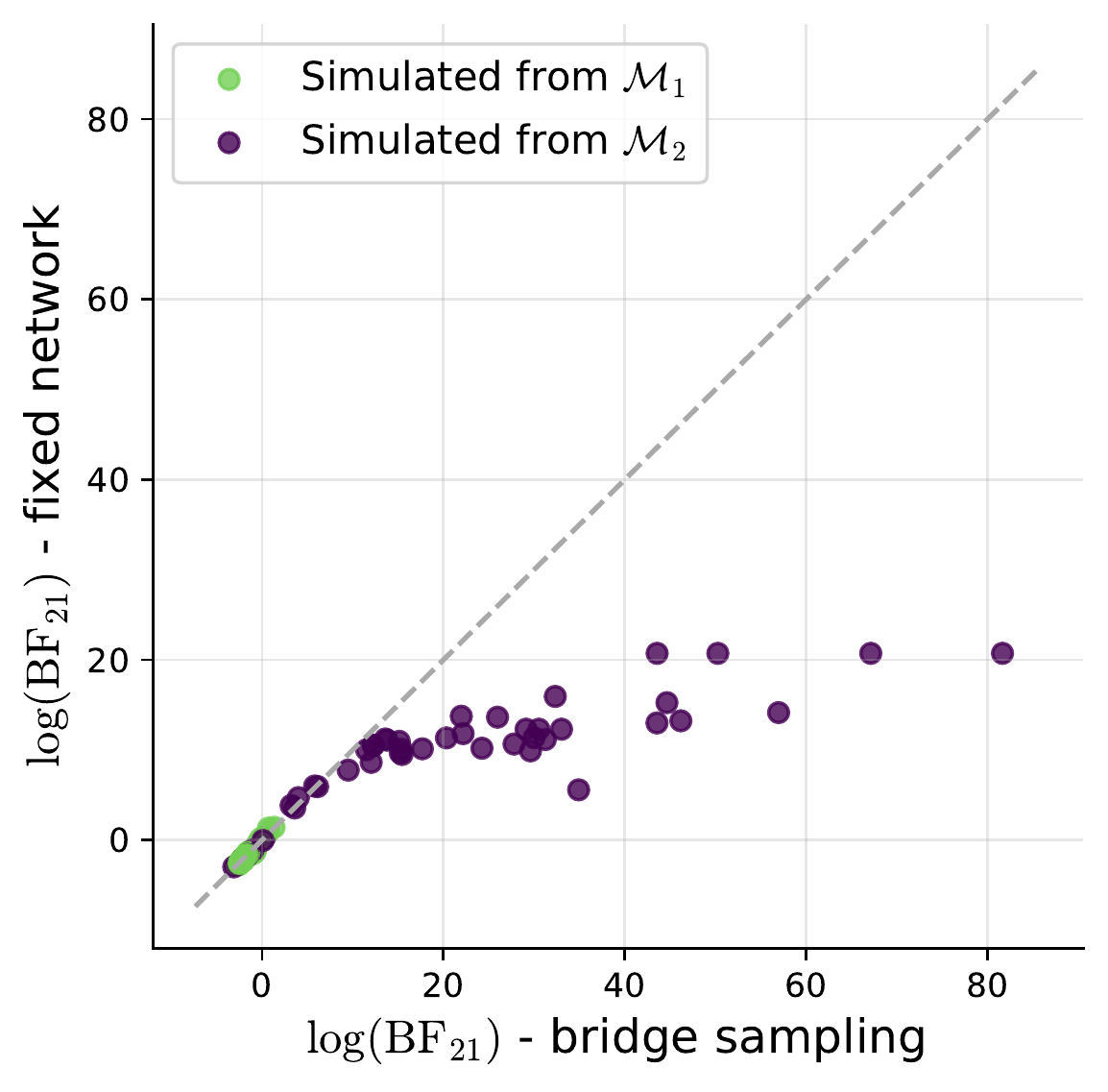}
    \end{subfigure}
    \begin{subfigure}{0.49\textwidth}
     \includegraphics[width=\textwidth]{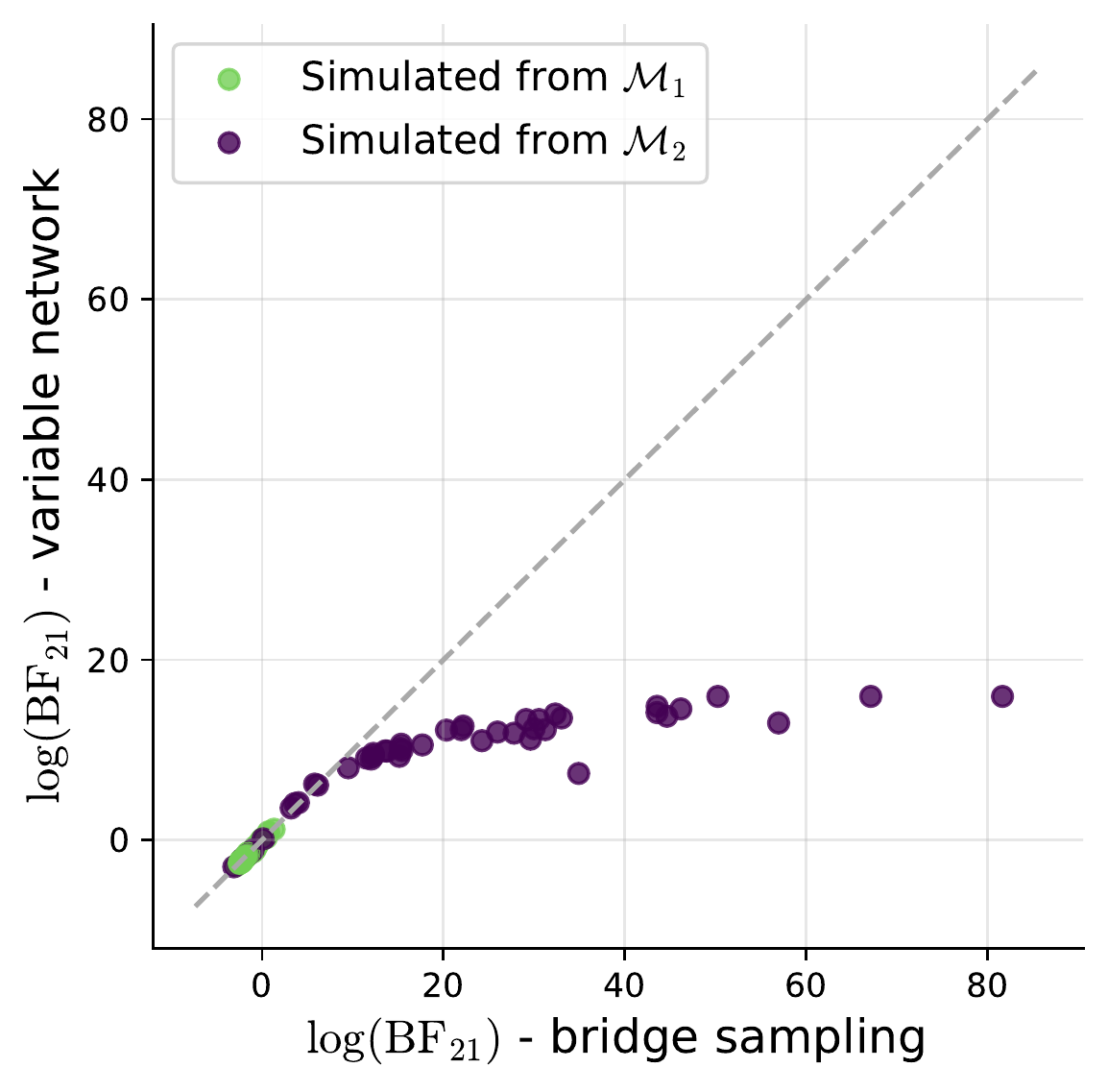}
    \end{subfigure}
    \caption{Validation study 1: Full comparison results for the log Bayes factors (all $100$ test data sets).}
    \label{fig:BS_vs_NN_BFs_full}
\end{figure*}

\autoref{fig:BS_vs_NN_BFs_full} displays the log BFs approximated by bridge sampling and the neural network variants for all $100$ test data sets, including those $27$ data sets for which bridge sampling approximated a BF $>1,000,000$ and that were therefore excluded in \autoref{fig:BS_vs_NN_convergence} for visibility purposes.

\section{Validation Study 2 Details}
\label{sec:val2_details}

\begin{figure*}[h]
    \centering
    \includegraphics[width=0.75\textwidth]{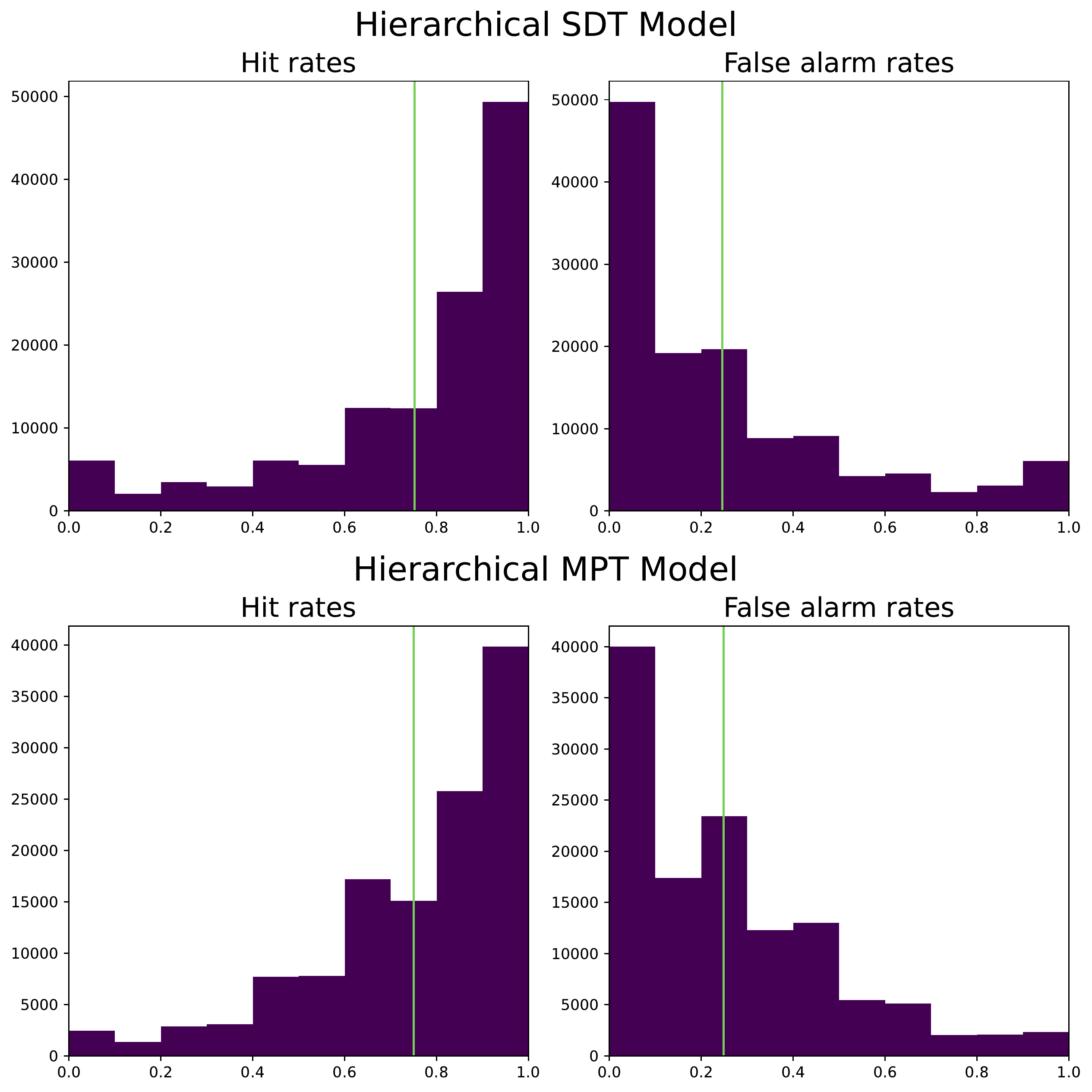}
    \caption{Validation study 2: Prior predictive checks for the SDT and the MPT model. The green vertical lines indicate the mean.}
    \label{fig:pp_MPTSDT}
\end{figure*}

Here, we provide details on our model specifications and prior choices.
We reformulate the observation-level structure of the MPT model as a binomial instead of a multinomial process to obtain identical response generation implementations for both models
\begin{align}
x_{mn}^{h} &\sim \text{Bernoulli}(h_m) \text{ for } n=1,\dots,N_m\\
x_{mn}^{f} &\sim \text{Bernoulli}(f_m) \text{ for } n=1,\dots,N_m,
\end{align}
where $h_m$ denotes the probability of detecting an old item as old ("hit") and $f_m$ denotes the probability of detecting a new item as old ("false alarm").
The generating processes of these probabilities with our distributional choices are described in Tables ~\ref{tab:sdt_hyperpriors} and ~\ref{tab:sdt_grouppriors} for the SDT model and Tables ~\ref{tab:mpt_hyperpriors} and ~\ref{tab:mpt_grouppriors} for the MPT models.
\autoref{fig:pp_MPTSDT} shows the prior predictive patterns of hit rates and false alarm rates arising from $5,000$ simulated data sets for each model.

\begin{figure*}[h]
    \begin{center}
    \includegraphics[width=0.5\textwidth, trim=1cm 0cm 0cm 0cm]{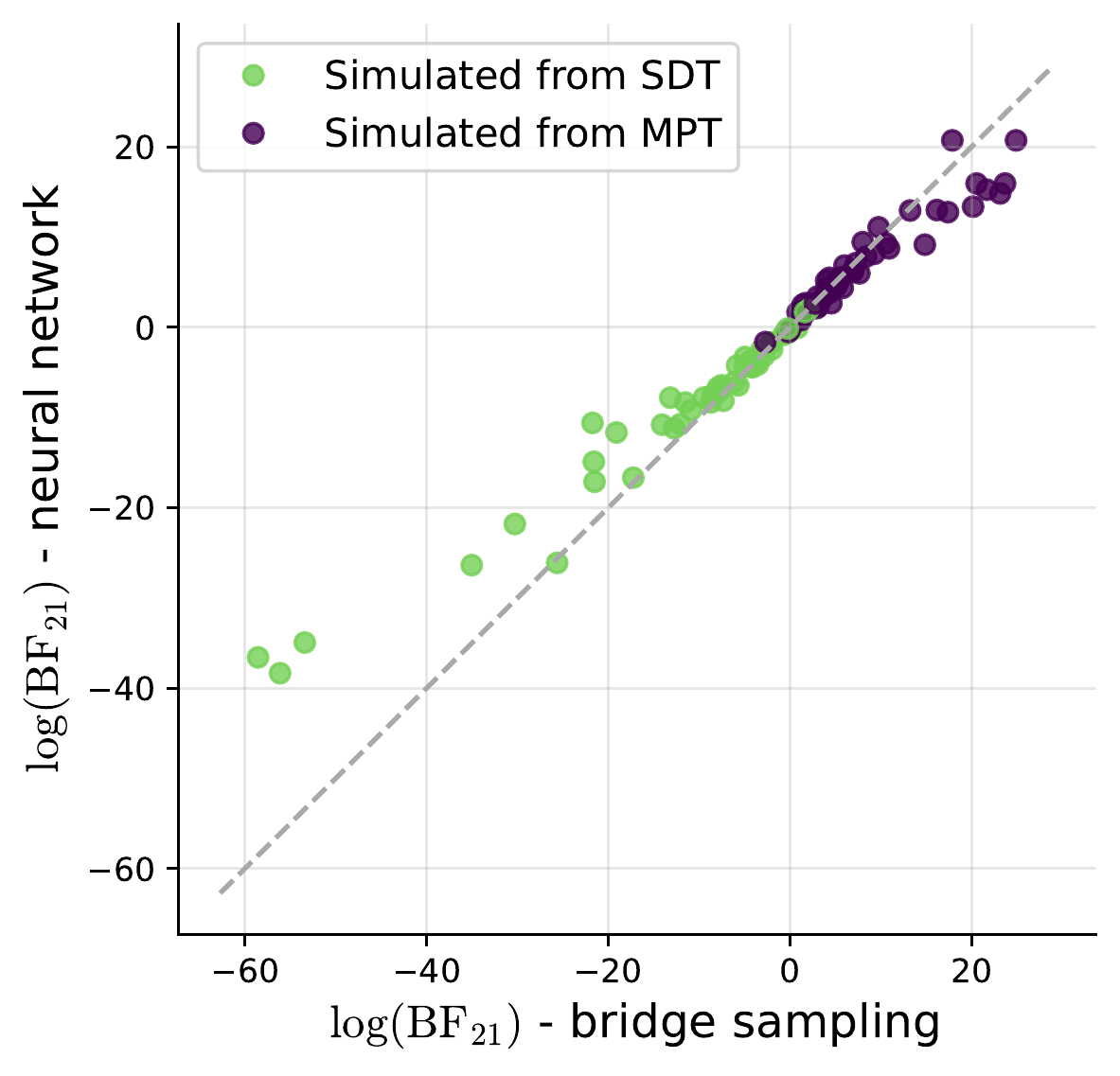}
    \caption{Validation study 2: Full comparison results for the log Bayes factors (all 100 test data sets).}
    \label{fig:BS_vs_NN_SDTMPT_BFs_full}
    \end{center}
\end{figure*}

\autoref{fig:BS_vs_NN_SDTMPT_BFs_full} presents the log BFs approximated by bridge sampling and the neural network, showing slight discrepancies in areas of extreme evidence. 
In contrast to the nested models in Validation Study 1, the SDT and MPT models being non-nested allows for extreme evidence for both models.

\begin{table*}[h]
\centering
\caption{Validation study 2: Hyperprior distributions of the SDT model.}
\label{tab:sdt_hyperpriors}
\begin{tabular}{@{}lcl@{}}
\toprule
Parameter & Symbol & Prior distribution \\
\midrule
\multirow{2}{*}[-2pt]{Probit-transformed hit probability} & $\mu_{h'}$ & Normal$(1, 0.5)$ \\ \cmidrule(l){2-3} 
                                            &  $\sigma_{h'}$ & Gamma$(1, 1)$ \\ \cmidrule{1-3}
\multirow{2}{*}[-2pt]{Probit-transformed false alarm probability} & $\mu_{f'}$ & Normal$(-1, 0.5)$ \\ \cmidrule(l){2-3} 
                                               & $\sigma_{f'}$ &  Gamma$(1, 1)$ \\
\bottomrule
\end{tabular}
\end{table*}

\begin{table*}[h]
\centering
\caption{Validation study 2: Group-level prior distributions and transformations of the SDT model.}
\label{tab:sdt_grouppriors}
\begin{tabular}{@{}lcl@{}}
\toprule
Parameter & Symbol & Prior distribution / transformation \\
\midrule
Probit-transformed hit probability & $h_m'$ & Normal$(\mu_{h'}, \sigma_{h'})$ \\
Probit-transformed false alarm probability & $f_m'$ & Normal$(\mu_{f'}, \sigma_{f'})$ \\
Hit probability & $h_m$ & $\Phi(h_m')$ \\
False alarm probability & $f_m$ & $\Phi(f_m')$ \\
\bottomrule
\end{tabular}
\end{table*}

\begin{table*}[h]
\centering
\caption{Validation study 2: Hyperprior distributions and transformations of the MPT model.}
\label{tab:mpt_hyperpriors}
\begin{tabular}{@{}lcl@{}}
\toprule
Parameter & Symbol & Prior distribution / transformation \\
\midrule
Probit-transformed recognition probability & $h_{d'}$ & Normal$(0, 0.25)$ \\ \cmidrule{1-3}
Probit-transformed guessing probability & $h_{g'}$ & Normal$(0, 0.25)$ \\ \cmidrule{1-3}
\multirow{4}{*}[-10pt]{Covariance matrix} & $\lambda_{d'}$ & Uniform$(0, 2)$ \\ \cmidrule(l){2-3} 
                                            &  $\lambda_{g'}$ & Uniform$(0, 2)$ \\ \cmidrule(l){2-3} 
                                            &  $Q$ & InvWishart$(3, \mathbb{I})$ \\ \cmidrule(l){2-3} 
                                            &  $\Sigma$ & Diag$(\lambda_{d'}, \lambda_{g'})$ $Q$ Diag$(\lambda_{d'}, \lambda_{g'})$ \\
\bottomrule
\end{tabular}
\end{table*}

\begin{table*}[h]
\centering
\caption{Validation study 2: Group-level prior distributions and transformations of the MPT model.}
\label{tab:mpt_grouppriors}
\begin{tabular}{@{}lcl@{}}
\toprule
Parameter & Symbol & Prior distribution / transformation \\
\midrule
Probit-transformed recognition probability & $d_m'$ & \multirow{2}{*}{Normal$ \left( \begin{bmatrix} \mu_{d'} \\ \mu_{g'} \end{bmatrix}, \Sigma \right)$} \\
Probit-transformed guessing probability & $g_m'$ & \\
Recognition probability & $d_m$ & $\Phi(d_m')$ \\
Guessing probability & $g_m$ & $\Phi(g_m')$ \\
Hit probability & $h_m$ & $d_m + (1-d_m)*g_m$ \\
False alarm probability & $f_m$ & $(1-d_m)*g_m$ \\
\bottomrule
\end{tabular}
\end{table*}

\section{Application Details}
\label{sec:application_details}

\subsection{Parameter Priors and Prior Predictive Checks}
\label{subsec:app_priors}

We base our priors upon the comprehensive collection of diffusion model parameter estimates by \cite{tran2021diffusion}.
For the Lévy flight models, \(\modeltwo\) and \(\modelfour\), we inform the prior on the additional $\alpha$ parameter by the estimates for comparable tasks (those completed under speed instructions) in \cite{voss2019sequential}.
For the inter-trial variability parameters included in \(\modelthree\) and \(\modelfour\), we follow the non-hierarchical priors that \cite{wiecki2013hddm} suggest to use in hierarchical drift-diffusion models, but choose a non-pooling approach with individual parameters instead of a complete-pooling approach.
Table~\ref{tab:diff_hyperpriors} contains the hyperprior choices and Table~\ref{tab:diff_grouppriors} the group-level priors.

\begin{figure*}[h]
    \centering
    \includegraphics[width=\textwidth]{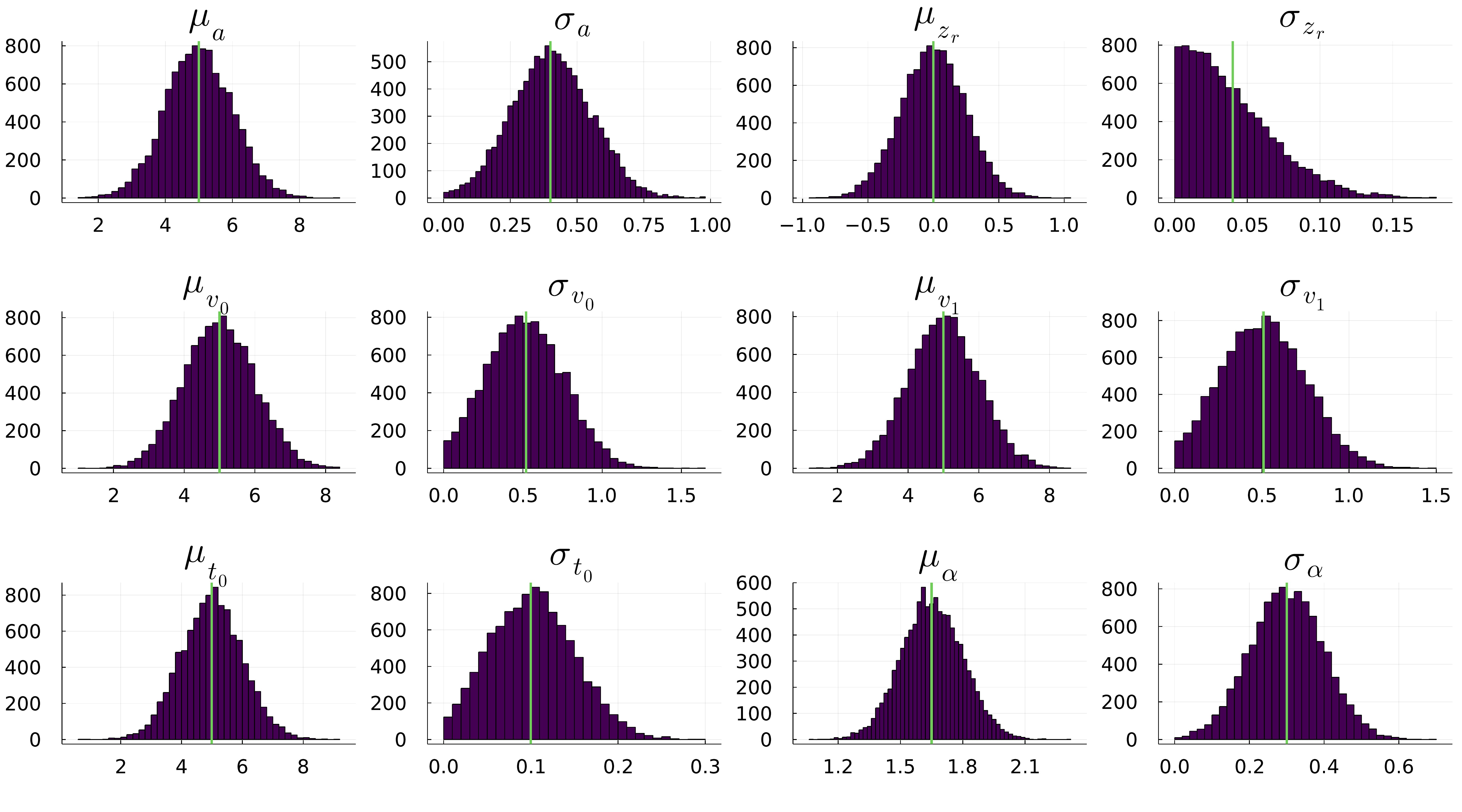}
    \caption{Real-data application: Prior predictive checks for the hyperpriors in the comparison of evidence accumulation models. The green vertical lines indicate the mean.}
    \label{fig:diff_pp_hyperpriors}
\end{figure*}

\begin{figure*}[h]
    \centering
    \includegraphics[width=0.75\textwidth]{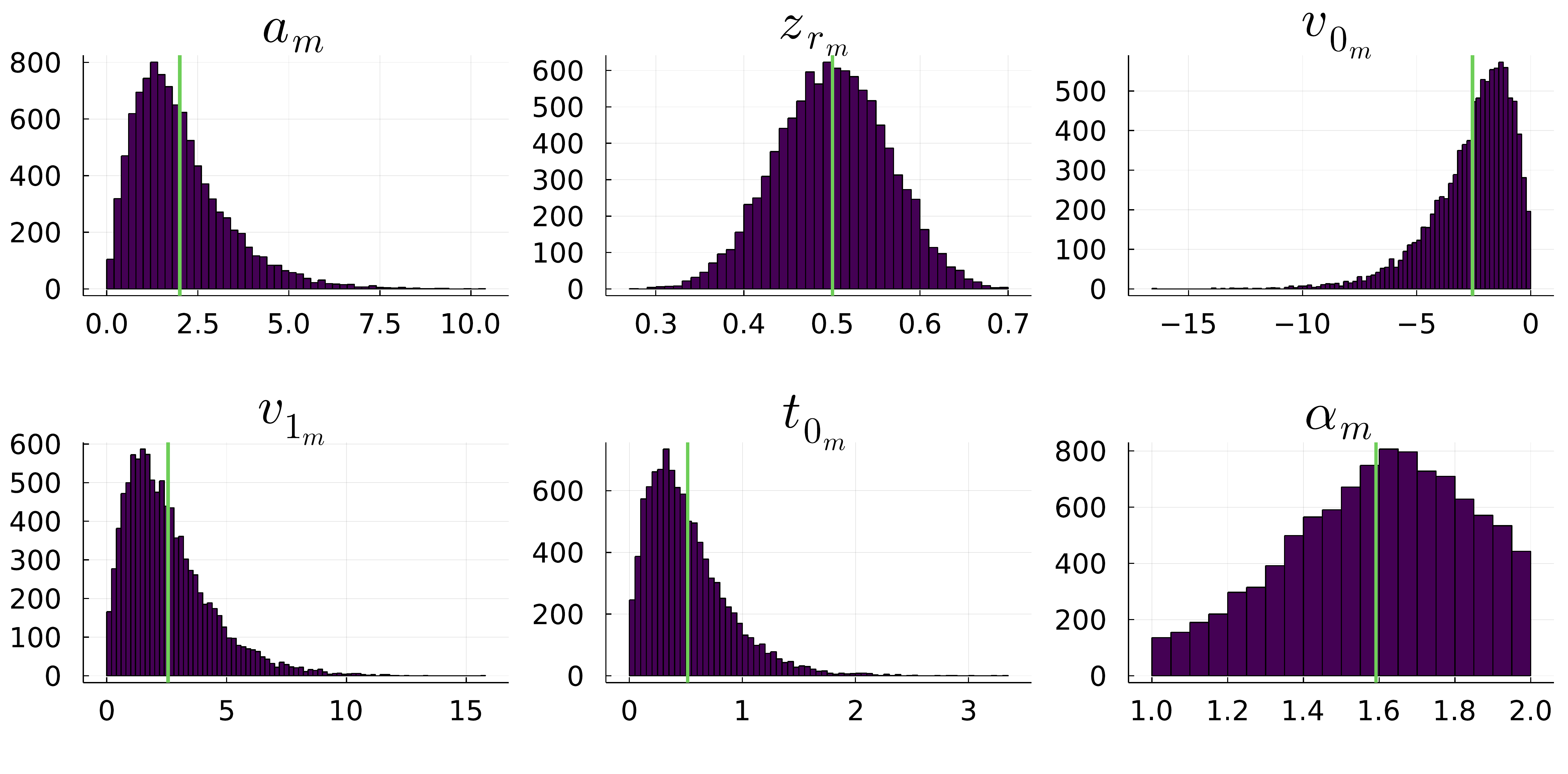}
    \caption{Real-data application: Prior predictive checks for the hierarchical group-level priors in the comparison of evidence accumulation models. The green vertical lines indicate the mean.}
    \label{fig:diff_pp_hierarchicalgroup}
\end{figure*}

\begin{figure*}[h]
    \centering
    \includegraphics[width=0.75\textwidth]{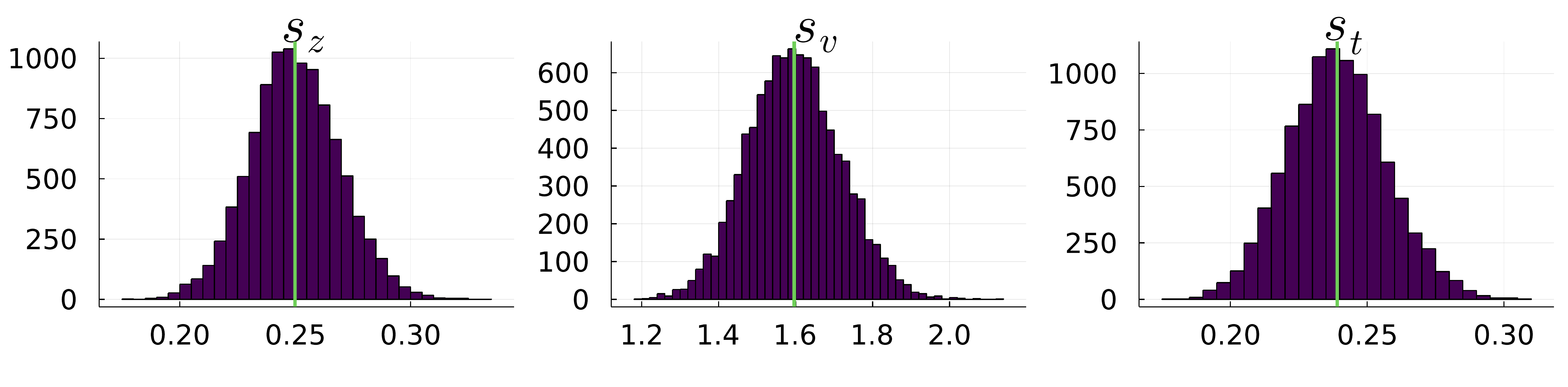}
    \caption{Real-data application: Prior predictive checks for the non-hierarchical group-level priors in the comparison of evidence accumulation models. The green vertical lines indicate the mean.}
    \label{fig:diff_pp_nonhierarchicalgroup}
\end{figure*}

To ensure that the informed priors for our HMs accurately reflect prior knowledge at both levels, we conduct prior predictive checks based on $10,000$ simulations (displayed in Figures~\ref{fig:diff_pp_hyperpriors}, \ref{fig:diff_pp_hierarchicalgroup} and \ref{fig:diff_pp_nonhierarchicalgroup}). 

\begin{table*}[h]
\centering
\caption{Real-data application: Hyperprior distributions of the evidence accumulation models.}
\label{tab:diff_hyperpriors}
\begin{tabular}{@{}lcl@{}}
\toprule
Parameter & Symbol & Prior distribution \\
\midrule
\multirow{2}{*}[-2pt]{Threshold separation} & $\mu_a$ & Normal$(5, 1)$ \\ \cmidrule(l){2-3} 
                                            &  $\sigma_a$ & Normal$_+(0.4, 0.15)$ \\ \cmidrule{1-3}
\multirow{2}{*}[-2pt]{Relative starting point} & $\mu_{zr}$ & Normal$(0, 0.25)$ \\ \cmidrule(l){2-3} 
                                               & $\sigma_{zr}$ &  Normal$_+(0, 0.05)$ \\ \cmidrule{1-3}
\multirow{2}{*}[-2pt]{Drift rate for blue/non-word stimuli} & $\mu_{v_0}$ & Normal$(5, 1)$ \\ \cmidrule(l){2-3} 
                                                            & $\sigma_{v_0}$ &  Normal$_+(0.5, 0.25)$ \\ \cmidrule{1-3}
\multirow{2}{*}[-2pt]{Drift rate for orange/word stimuli} &  $\mu_{v_1}$ & Normal$(5, 1)$ \\ \cmidrule(l){2-3} 
                                                          &  $\sigma_{v_1}$ &  Normal$_+(0.5, 0.25)$ \\ \cmidrule{1-3}
\multirow{2}{*}[-2pt]{Non-decision time} & $\mu_{t_0}$ & Normal$(5, 1)$ \\ \cmidrule(l){2-3} 
                                           & $\sigma_{t_0}$ &  Normal$_+(0.1, 0.05)$ \\ \cmidrule{1-3}
\multirow{2}{*}[-2pt]{Stability parameter of the noise distribution} & $\mu_\alpha$ & Normal($1.65, 0.15$) \\ \cmidrule(l){2-3} 
                                                                     & $\sigma_\alpha$ &  Normal$_+(0.3, 0.1)$ \\
\bottomrule
\end{tabular}
\footnotetext{\textit{Note.} Normal$_+(\cdot)$ denotes a zero-truncated normal distribution that only allows for positive values.}
\end{table*}

\begin{table*}[h]
\centering
\caption{Real-data application: Group-level prior distributions of the evidence accumulation models.}
\label{tab:diff_grouppriors}
\begin{tabular}{@{}lcl@{}}
\toprule
Parameter & Symbol & Prior distribution \\
\midrule
Threshold separation & $a_m$ & Gamma$(\mu_a, \sigma_a)$ \\
Relative starting point & $zr_m$ & invlogit(Normal$(\mu_{zr}, \sigma_{zr}))$ \\
Drift rate for blue/non-word stimuli & $v_{0_m}$ & -Gamma$(\mu_{v_0}, \sigma_{v_0})$ \\
Drift rate for orange/word stimuli & $v_{1_m}$ & Gamma$(\mu_{v_1}, \sigma_{v_1})$ \\
Non-decision time & $t_{0_m}$ & Gamma$(\mu_{t_0}, \sigma_{t_0})$ \\
Stability parameter of the noise distribution & $\alpha_m$ & TruncatedNormal$(\mu_\alpha, \sigma_\alpha, 1, 2)$ \\
Inter-trial variability of starting point & $s_{z_m}$ & Beta$(1, 3)$ \\
Inter-trial variability of drift & $s_{v_m}$ & Normal$_+(0, 2)$ \\ 
Inter-trial variability of non-decision time & $s_{t_m}$ & Normal$_+(0, 0.3)$ \\ 
\bottomrule
\end{tabular}
\footnotetext{\textit{Note.} Normal$_+(\cdot)$ denotes a zero-truncated normal distribution that only allows for positive values. TruncatedNormal$(\cdot)$ denotes a truncated normal distribution with the lower and upper limit given by the last two values.}
\end{table*}

\subsection{Robustness Against Artificial Noise}
\label{subsec:app_robustness}

\begin{figure*}[h]
    \centering
    \includegraphics[width=0.5\textwidth]{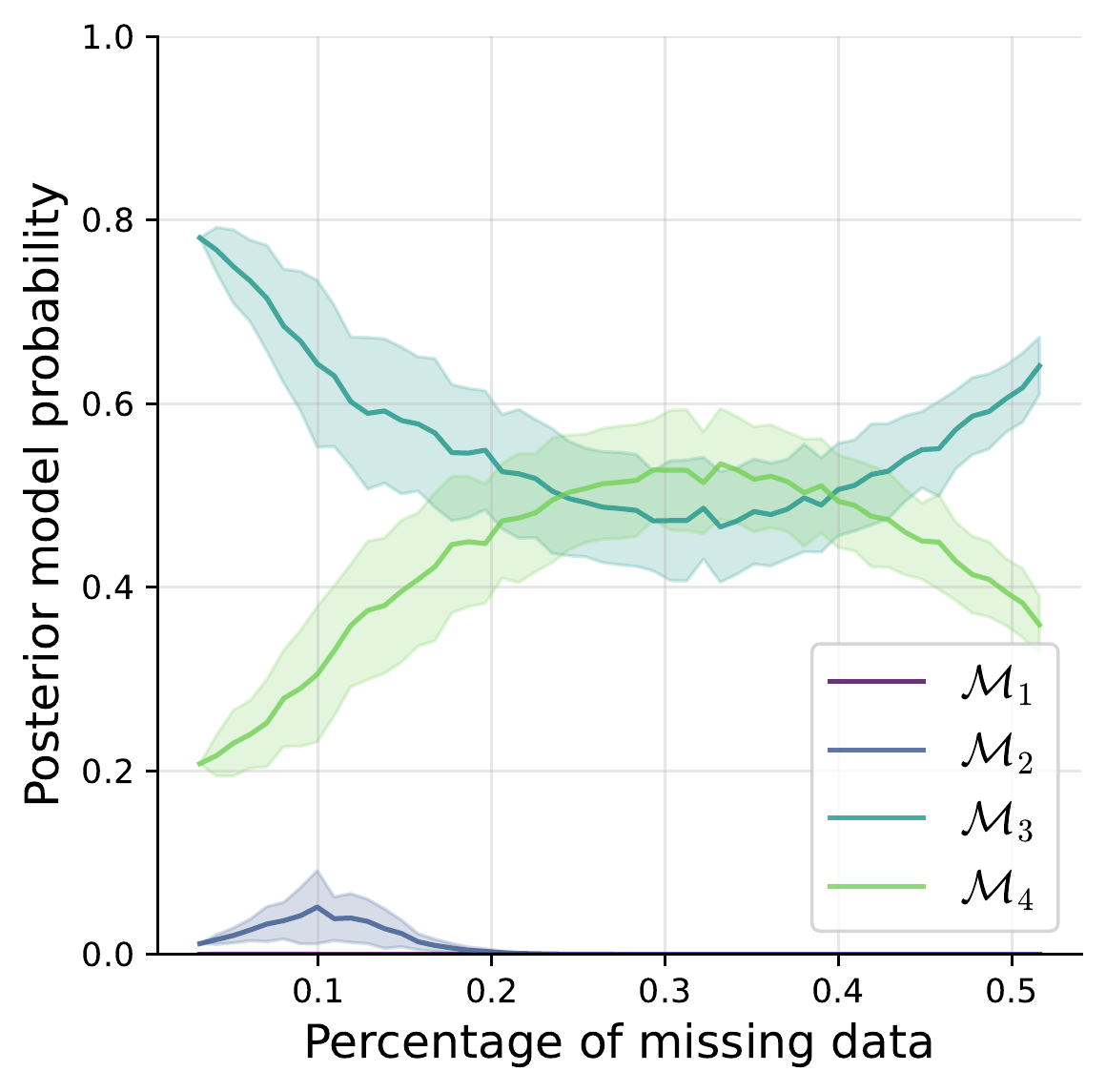}
    \caption{Real-data application: Robustness of the model comparison results against increasing amounts of artificially injected random noise. The lines represent the average probabilities of $100$ repetitions per percentage step (in each repetition masking a random subset of the empirical data), whereas the shaded areas indicate the standard deviation between these repetitions.}
    \label{fig:noise_robustness}
\end{figure*}

Here, we inspect the stability of our neural network against additional noise injection.
\autoref{fig:noise_robustness} displays the model comparison results as increasing percentages of trials per participant are artificially masked as missing. 
We repeat the random masking of trials $100$ times per percentage step to assess the sensitivity of the results to specific parts of the empirical data.
Consistent with our main results, there is a clear separation between low evidence for \(\modelone\) and \(\modeltwo\) and substantial evidence for \(\modelthree\) and \(\modelfour\) across all settings.
Despite our network being trained on the empirical amount of missing data, 3.17\% over both tasks, we observe rank stability of the model comparison results up until 25\% missing data per participant.

\end{appendices}

\end{document}

%% file: preamble.tex
\usepackage[backend=biber,
style=apa,
maxcitenames=2,
maxbibnames=100, 
language=english,
doi=false,
isbn=false,
url=false,
uniquename=false
]{biblatex} 
\DeclareLanguageMapping{english}{english-apa} 
\usepackage{subcaption}
\usepackage{placeins}
\usepackage{mathtools}
\usepackage{amssymb}
\usepackage{amsmath}
\usepackage{amsfonts}   
\usepackage{fancyhdr}
\usepackage{times}
\usepackage{amsthm}
\usepackage{siunitx}
\usepackage{nicefrac}
\usepackage{booktabs}
\usepackage{multirow}
\usepackage{appendix}

\usepackage{todonotes}
\usepackage{algorithm}
\usepackage{algorithmic}

\newcommand{\bs}{\boldsymbol}
\newcommand{\xb}{\bs{x}}

\newcommand{\etab}{\bs{\eta}}
\newcommand{\zb}{\bs{z}}
\newcommand{\thetab}{\bs{\theta}}
\newcommand{\phib}{\bs{\phi}}
\newcommand{\pib}{\bs{\pi}}

\newcommand{\given}{\,|\,}
\newcommand{\model}{\mathcal{M}_j}
\newcommand{\modelk}{\mathcal{M}_k}
\newcommand{\modelone}{\mathcal{M}_1}
\newcommand{\modeltwo}{\mathcal{M}_2}
\newcommand{\modelthree}{\mathcal{M}_3}
\newcommand{\modelfour}{\mathcal{M}_4}

\newcommand{\lik}{p\left(\xb\given\thetab_j, \model\right)}
\newcommand{\prior}{p\left(\thetab_j\given\model\right)}
\newcommand{\joint}{p\left(\thetab_j, \xb\given\model\right)}

\usepackage{float}
\usepackage{enumitem}

\usepackage{xcolor}
\definecolor{darkblue}{RGB}{0,0,128}
\definecolor{darkgreen}{RGB}{0, 128, 0}
\definecolor{darkred}{RGB}{128, 0, 0}
\definecolor{black}{RGB}{0, 0, 0}
\definecolor{errorcolor}{HTML}{481567}
\definecolor{viridisgreen}{HTML}{55C667}

\usepackage{yfonts}